\documentclass[10pt]{article} 
\usepackage[preprint]{tmlr}

\usepackage{amsmath,amsfonts,bm}

\def\eqref#1{equation~\ref{#1}}

\def\1{\bm{1}}

\DeclareMathAlphabet{\mathsfit}{\encodingdefault}{\sfdefault}{m}{sl}
\SetMathAlphabet{\mathsfit}{bold}{\encodingdefault}{\sfdefault}{bx}{n}

\usepackage{hyperref}
\usepackage[table,xcdraw]{xcolor}
\hypersetup{colorlinks=true,linkcolor=blue!55!black,citecolor=blue!55!black,urlcolor=blue!55!black}
\usepackage{multirow}
\usepackage{url}            
\usepackage{booktabs}       
\usepackage{amsfonts}       
\usepackage{nicefrac}       
\usepackage{microtype}      
\usepackage{amsmath,graphicx}
\usepackage{subcaption}
\usepackage{amssymb}
\usepackage{xspace}
\usepackage{enumitem}
\usepackage[accsupp]{axessibility} 
\usepackage{breqn}
\usepackage{minted}
\usepackage{algorithmic}
\usepackage{algorithm2e}
\usepackage{bbm}

\title{Stabilizing Camera-Controlled Novel View Synthesis at Inference Time\thanks{Project page at \url{https://prajwalsingh.github.io/CamTrolpp/} and code at \url{https://github.com/prajwalsingh/CamTrolpp}.}}

\author{\name Prajwal Singh, Arjun Badola, Seema Kumari, Hajime Nagahara$^{\dagger}$, Shanmuganathan Raman \\
      \addr CVIG Lab and ISLab, D3 Center$^{\dagger}$ \\
      \addr IIT Gandhinagar and Osaka University$^{\dagger}$}

\def\month{MM}  
\def\year{YYYY} 
\def\openreview{\url{https://openreview.net/forum?id=XXXX}} 

\begin{document}
\maketitle

\begin{abstract}
\vspace{-1.0em}
Training-free, camera-controlled novel view synthesis from a single image using pre-trained video diffusion models often becomes unstable under large camera motion and long generation horizons. Existing approaches commonly combine several inference-time components, making it unclear which design choices are most important for stability. This work identifies the amount of camera motion handled by each diffusion pass as the main factor. CamTrol++ splits the camera path into passes that each cover at most a fixed rotation and generates them one after another, so that every pass stays within the range of motion the backbone handles well. A matched study, in which every setting is scored on the same poses of the same $60^\circ$ orbit, shows that geometric error grows steadily with the arc per pass, while a single SVD pass both loses geometry and tends to freeze the camera. Under one shared camera path, CamTrol++ achieves the lowest epipolar error and the best held-out 3D Gaussian Splatting reconstruction among CamTrol, NVS-Solver and WAVE, and reaches the quality of CamTrol with the same pass structure at least $3.2\times$ faster. On real RealEstate10K motion it matches CamTrol at half the run time (NVS-Solver is more faithful but $8\times$ slower). The same procedure also improves 3D reconstruction with a second video backbone, LTX-Video. Geometry-constrained spatial attention, color anchoring at pass boundaries and a registration-free warping pipeline that makes warping $6.1\times$ faster complete the framework. The method remains effective for 56-frame generation and under controlled depth corruption. These results show that careful control of camera motion at inference time can substantially improve the stability of camera-controlled novel view synthesis.

\vspace{-1em}
\end{abstract}

\section{Introduction}
\label{sec:intro}

\begin{figure}[!t]
\centering 
\includegraphics[width=0.92\linewidth]{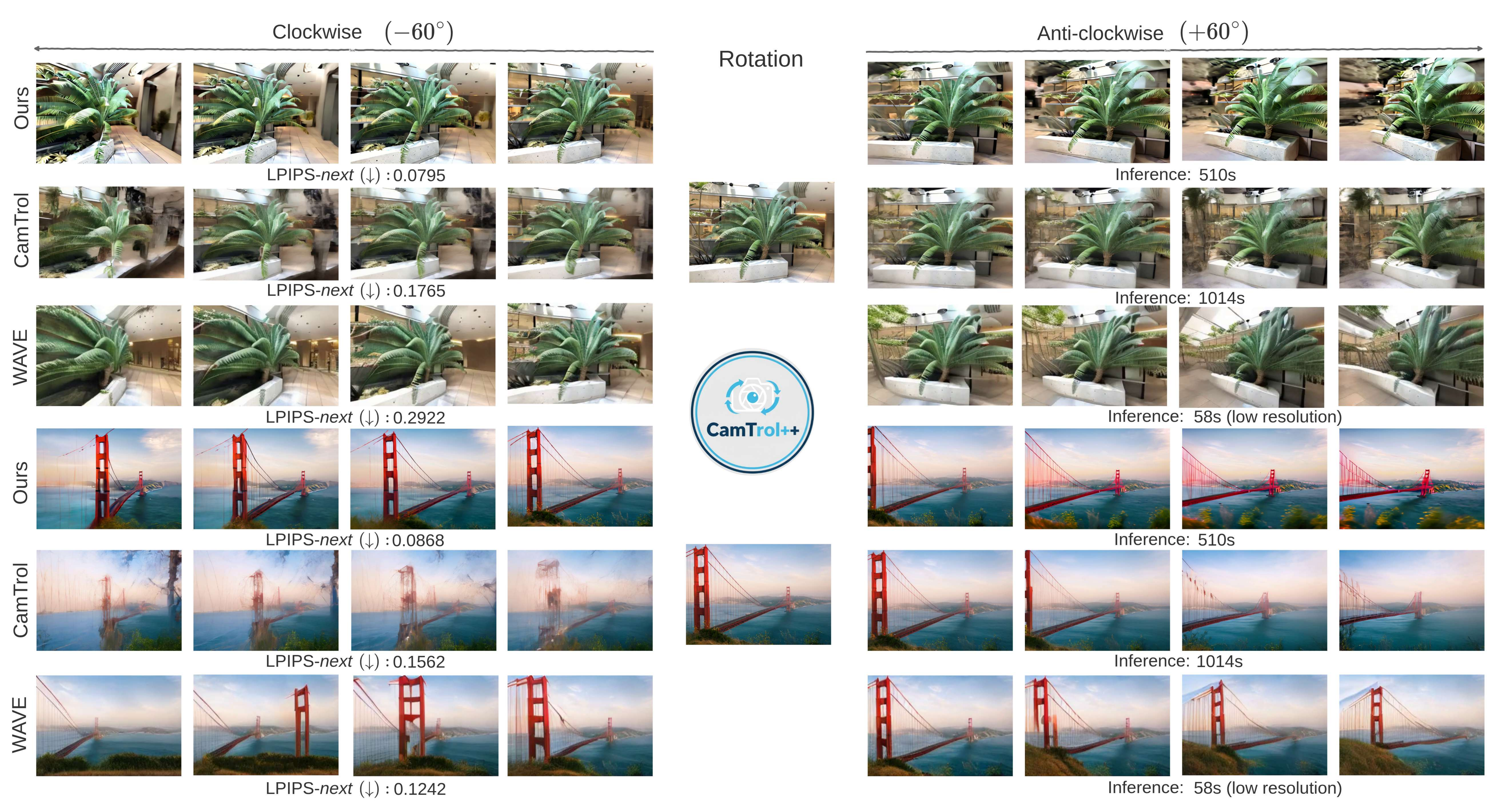}\vspace{-0.7em}
\captionof{figure}{\textbf{Large Camera Motion.} CamTrol++ is compared with representative training-free baselines (CamTrol~\citep{hou2024training}, WAVE~\citep{park2025wave}) on a challenging $\pm60^{\circ}$ rotation. Baselines fail to maintain coherence, resulting in significant distortion and high temporal flicker. CamTrol++ splits the large motion into small autoregressive steps, generating a stable, geometrically coherent sequence with $1.4$ to $3.7\times$ lower LPIPS-\textit{next}. Times are for this $56$-frame example (matched timings in Table~\ref{tab:merged_geometry_3dgs}).}
\label{fig:teaser}
\vspace{-1.0em}
\end{figure}

The availability of pre-trained diffusion models for images and videos has enabled applications ranging from virtual reality to content creation~\citep{poole2022dreamfusion,brooks2023instructpix2pix,liu2023zero1to3zeroshotimage3d,han2025zero4d}. Recent advances have made it possible to generate compositional and photorealistic images and videos in a training-free setting~\citep{hertz2022prompt,feng2022training,chefer2023attend,agarwal2023star,kim2024fifo,yang2024compositional,kwon2024tweediemix}. An important direction is camera and view control for diffusion-based image and video generation, where the goal is to synthesize multiple views of a scene from a single input image~\citep{sun2024dimensionx,cui2025uniview}, alongside training-free work on consistent 3D scene editing~\citep{gomel2024diffusion} and geometric image editing~\citep{zhu2025training}, useful for 3D content creation and scene exploration.

Existing approaches to novel view synthesis can be broadly divided into two categories. Per-scene optimization methods, such as Neural Radiance Fields (NeRF)~\citep{mildenhall2021nerf} and 3D Gaussian Splatting~\citep{kerbl20233d}, achieve high-quality reconstruction but require multi-view data and scene-specific optimization. Zero-shot methods, such as Zero-1-to-3~\citep{liu2023zero1to3zeroshotimage3d}, synthesize new viewpoints from a single image but often struggle with scene consistency and long-range stability. More recently, training-free methods have combined depth-based warping with diffusion refinement. CamTrol~\citep{hou2024training} uses a warp-and-inpaint framework with Stable Video Diffusion (SVD)~\citep{blattmann2023stable}, while WAVE~\citep{park2025wave} generates all views jointly with an image diffusion model, linking them through warp-guided attention. NVS-Solver~\citep{you2025nvssolver} instead guides the denoising of a video model with warped views at every sampling step. Although effective for short sequences, these methods degrade under large camera motions and longer trajectories.

The main challenge is error accumulation during autoregressive generation. Each generated frame is used to generate the next, so small geometric and appearance errors can accumulate over time. Large camera motions make this problem more severe by introducing larger reprojection errors and disoccluded regions within each generation step. This leads to geometric distortion, perspective drift, temporal flicker, and appearance drift, as shown in Figure~\ref{fig:teaser}. These issues are addressed by \textit{CamTrol++}, a training-free framework that stabilizes camera-controlled novel view synthesis at inference time without modifying the diffusion backbone.

The key idea is to bound the camera motion that each diffusion pass has to handle. CamTrol++ divides the camera path into passes that each rotate the camera by at most $15^\circ$ and generates them one after another, rebuilding the point cloud from the last generated frame at every pass boundary. Short paths therefore use a single pass, and long paths are split into as many passes as the bound requires. Geometry-constrained spatial attention and low-frequency appearance anchoring at pass boundaries refine geometry and color, and a registration-free warping pipeline replaces the costly point cloud registration of CamTrol~\citep{hou2024training}. A matched step-size study, in which every arc per pass is scored on the same poses of the same orbit, shows that geometric error grows steadily with the arc, and that a single SVD pass both loses geometry and freezes large parts of the image. Splitting also improves 3D reconstruction and helps a second backbone, LTX-Video, whose best arc per pass is larger ($20^\circ$ to $30^\circ$). The main contributions are as follows.

\begin{figure*}[!t]
\centering
\begin{minipage}[b]{0.95\linewidth}
  \centering
  \centerline{\includegraphics[width=\linewidth]{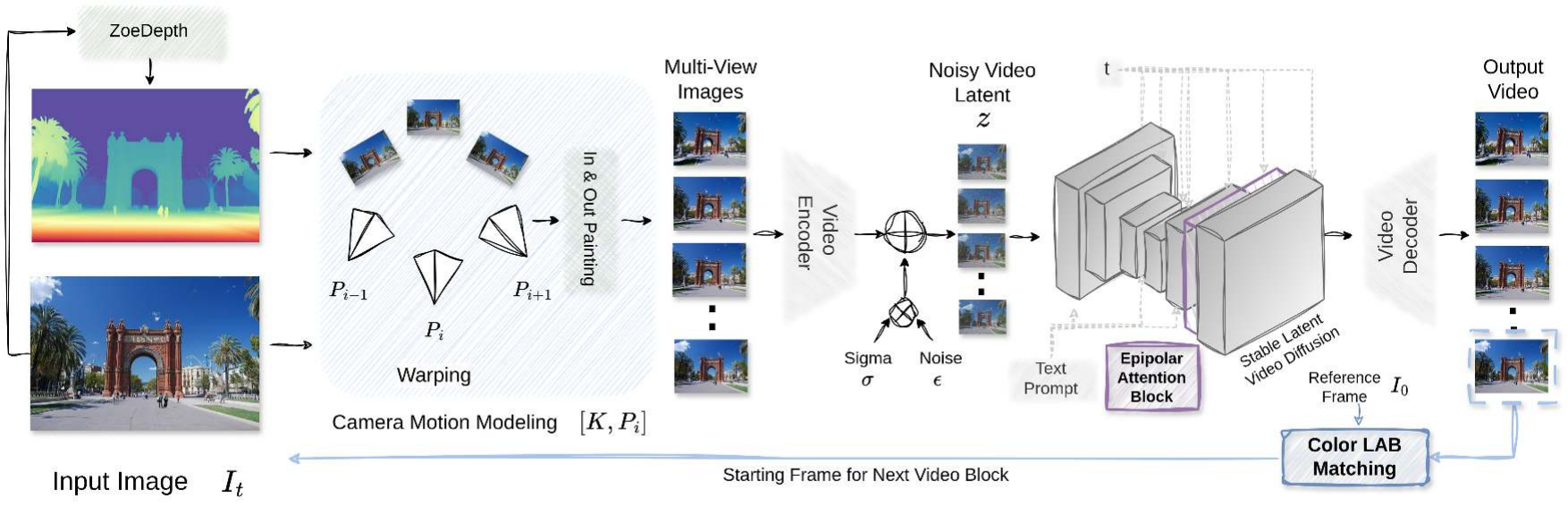}}\vspace{-1.3em}
\end{minipage}
\captionof{figure}{\textbf{Overview of the CamTrol++ Framework.} Given the input frame $I_t$ of the current pass ($I_{\text{input}}$ in Section~\ref{sec:method}) and its depth, the Camera Motion Modeling block generates a warped multi-view sequence along a small camera sub-path. The sequence is processed by Stable Video Diffusion with \textit{Epipolar Attention} for geometric consistency and \textit{Color LAB Matching} for appearance consistency. The corrected frame is reused as input for the next autoregressive pass, enabling efficient long-horizon generation without point cloud registration.}
\label{fig:figure_2_framework}
\vspace{-1.0em}
\end{figure*}

\begin{itemize}
    \item A training-free framework that stabilizes camera-controlled novel view synthesis by bounding the camera rotation per diffusion pass, without retraining or modifying the backbone. On a $60^\circ$ orbit it achieves the lowest epipolar error and the best held-out 3D Gaussian Splatting reconstruction among CamTrol, NVS-Solver and WAVE.
    \item A matched step-size study that isolates the effect of the arc per pass. Every setting follows the same path and is scored on the same poses, which shows that geometric error grows steadily with the arc and that a single SVD pass tends to stall.
    \item Evidence that the procedure transfers. It improves held-out 3D reconstruction with LTX-Video~\citep{hacohen2024ltx} as a second backbone, and it keeps its advantage under structured depth failures such as flattened, blurred or wrongly scaled depth.
    \item A registration-free warping pipeline that makes warping $6.1\times$ faster and a full pass $3.2\times$ faster than CamTrol~\citep{hou2024training}, together with LAB anchoring at pass boundaries that removes most of the color drift across passes.
\end{itemize}

\section{Related Work}
\label{sec:relatedworks}

\vspace{1mm} \noindent \textbf{Per-Scene Optimization.} Early methods for novel view synthesis, such as Neural Radiance Fields (NeRF)~\citep{mildenhall2021nerf}, learn volumetric radiance fields per scene through gradient-based optimization. Extensions like Mip-NeRF~\citep{barron2021mip} and Instant-NGP~\citep{muller2022instant} improve speed and fidelity but remain scene-specific. Similarly, 3D Gaussian Splatting (3DGS)~\citep{kerbl20233d} and dynamic variants~\citep{Lu_2024_CVPR} represent scenes with explicit Gaussian primitives, achieving photorealistic results through per-scene fitting. EgoGaussian~\citep{zhang2025egogaussian} extends this idea to dynamic and egocentric setups by jointly reconstructing backgrounds and moving objects. While these approaches produce geometrically consistent reconstructions, they require multi-view supervision and known camera poses, whereas the goal here is to generate multiple views of a scene from a single image. Recent works such as LucidDreamer~\citep{chung2023luciddreamer} and SpatialCrafter~\citep{zhang2025spatialcrafter} move toward sparse- or single-view 3D scene generation but rely on additional scene optimization or trained geometric components, in contrast to the inference-time approach studied here.

\vspace{1mm} \noindent \textbf{Zero-Shot Novel View Synthesis (NVS).} Zero-shot NVS methods generate novel viewpoints from a single image without per-scene training. Zero-1-to-3~\citep{liu2023zero1to3zeroshotimage3d} and Zero123++~\citep{shi2023zero123plus} use diffusion models for 3D-aware image generation, while ViewCrafter~\citep{yu2024viewcraftertamingvideodiffusion}, ViewDiff~\citep{hollein2024viewdiff}, and Consistent-1-to-3~\citep{ye2023consistent} improve geometric consistency through multi-view attention and geometry-aware diffusion. EpiDiff~\citep{huang2024epidiff} adds an epipolar-constrained attention block to a multi-view diffusion model, which is close in spirit to the geometry-constrained attention used here, although EpiDiff trains this block while CamTrol++ applies the constraint only at inference. ViVid-1-to-3~\citep{kwak2024vivid} combines a pre-trained view-conditioned diffusion model with a video diffusion model at sampling time, without training, for object-centric view synthesis, and GenWarp~\citep{seo2024genwarp} learns where to trust a depth warp inside a trained image model. \citet{elata2025novel} train a pixel-space diffusion model for novel view synthesis. WonderJourney~\citep{yu2024wonderjourney} uses language-guided scene expansion for perpetual 3D scene generation, while FantasyWorld~\citep{dai2025fantasyworld} adds a trained geometry branch to a frozen video model for joint video and 3D prediction. Stable Virtual Camera~\citep{zhou2025stable} focuses on camera-controlled diffusion. NVS-Solver~\citep{you2025nvssolver} injects warped views into every sampling step of a frozen video model, while WAVE~\citep{park2025wave} uses warp-guided attention and pose-aware noise initialization to improve consistency across novel views that an image diffusion model generates jointly. Concurrent with this work, NeoMap~\citep{li2026neomap} optimizes the initial noise of pre-trained video backbones, while WorldForge~\citep{song2025worldforge} refines each denoising step with motion guidance. These methods steer the camera within one denoising run and the native generation horizon of a pre-trained video backbone. The focus here is different, namely keeping views consistent across \emph{multiple} autoregressive windows. This setting matters for long or complex camera paths, where a single window is not enough. The study therefore targets multi-window stabilization and the role of each of its components.

\vspace{1mm} \noindent \textbf{Long-Horizon Video Generation and Camera Control.} This line of work extends pre-trained video models to longer or camera-controlled generation, either at inference time or with additional training. CamTrol~\citep{hou2024training} uses a warp-and-inpaint pipeline~\citep{fridman2023scenescape} with Stable Video Diffusion (SVD), enabling camera control without retraining. However, large camera motions can lead to accumulated geometric and perceptual errors. FreeLong~\citep{lu2024freelong} addresses long-video degradation through spectral feature blending in a training-free setting, while Rolling Forcing~\citep{liu2025rolling} reduces long-horizon error accumulation through joint denoising and context anchoring with additional training. \citet{sadat2024eliminating} reduce the oversaturation of high classifier-free guidance scales by reweighting the guidance update, without retraining. These methods improve long-horizon generation or sampling quality but do not specifically address geometric error accumulation in training-free camera-controlled novel view synthesis. Recent large video backbones such as LTX-Video~\citep{hacohen2024ltx} and Wan~\citep{wan2025wan} generate longer clips natively, yet their horizon is still finite. CamCo~\citep{xu2024camco} and CamI2V~\citep{zheng2024cami2v} explore camera-controllable image-to-video generation through fine-tuned or training-based camera pose conditioning.

Existing training-free approaches show that splitting the camera path and adding geometric or appearance constraints can improve the stability of autoregressive novel view synthesis. Large camera motions and long paths remain hard because errors accumulate across generation steps. This work studies stabilization across multiple autoregressive windows and measures the role of small-step camera decomposition and of the supporting geometric and appearance components. It also measures how the camera step size, the generation length and depth errors affect the practical behavior of the framework.

\section{Method}
\label{sec:method}

CamTrol++ is an inference-time stabilization framework for camera-controlled novel view synthesis. It runs on a pre-trained Stable Video Diffusion (SVD) backbone~\citep{blattmann2023stable} and needs no retraining or change to the weights. Given a single input image and a predefined camera trajectory, the framework generates multi-view video sequences in an autoregressive manner. Figure~\ref{fig:figure_2_framework} illustrates the overall pipeline (Algorithm~\ref{algo:main_algo}, Appendix~\ref{sec:supp_implementation}). The framework has one central mechanism, distortion-limited reprojection through small-step camera decomposition, together with three supporting components, which are efficient registration-free warping, low-frequency appearance anchoring, and geometry-constrained spatial attention.

\begin{figure}[!t]
\centering
\begin{minipage}[b]{0.95\linewidth}
  \centering
  \centerline{\includegraphics[width=\linewidth]{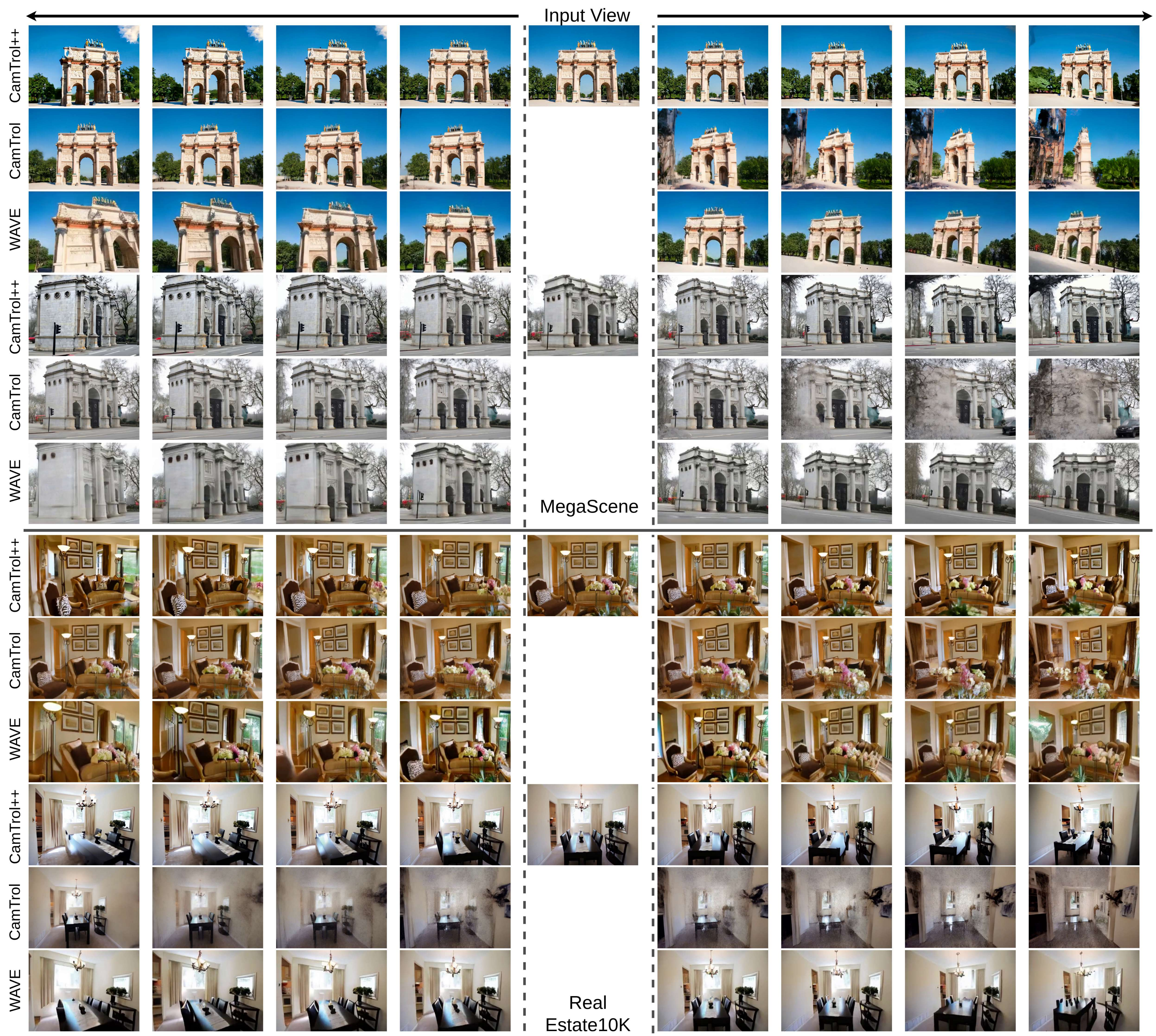}}
  \vspace{-0.5em}
\end{minipage}
\captionof{figure}{\textbf{Qualitative Comparison.} CamTrol++ produces sharp and consistent multi-view sequences, maintaining both visual quality and temporal stability. In contrast, the baseline CamTrol~\citep{hou2024training} generates distorted frames, while WAVE~\citep{park2025wave} introduces geometric distortions and artifacts. This comparison highlights the improved stability and overall quality of CamTrol++.}
\label{fig:figure_3_qualitative_comp}
\vspace{-1.0em}
\end{figure}

\label{sec:3.1} \vspace{1mm} \noindent \textbf{Camera-Consistent Distortion-Limited Reprojection.} A primary source of instability in prior training-free methods, such as CamTrol~\citep{hou2024training} and WAVE~\citep{park2025wave}, is large single-step camera motion. When wide rotations (e.g., $60^{\circ}$) or long translations are generated within a single $14$-frame chunk, severe geometric distortion can occur. The diffusion model must inpaint large disoccluded regions, often leading to structural artifacts and drift (Figures~\ref{fig:teaser},~\ref{fig:figure_3_qualitative_comp},~\ref{fig:figure_4_ablation}).

To reduce per-step geometric error, large camera motion is split into smaller segments. A continuous $M$-frame trajectory ($M=28$ or $56$, and $84$ in the long-horizon study, counting the shared frame of consecutive chunks twice) is first generated,\vspace{-0.3em}
\[
P_{world} = \{[R_0 \,|\, t_0], \dots, [R_{L-1} \,|\, t_{L-1}]\},
\]

\noindent
using a smooth procedural motion function, where each $[R\,|\,t]$ maps world coordinates to camera coordinates. Instead of synthesizing all frames in a single pass, the trajectory is divided into $N$-frame chunks (e.g., $N=14$) that are processed autoregressively, one diffusion pass per chunk. Consecutive chunks share one frame, since the last frame of a chunk is the first frame of the next. A chunk therefore advances the camera by $N-1=13$ pose intervals, and $K$ chunks produce $L=(N-1)K+1=13K+1$ distinct frames, one for each pose of $P_{world}$. For each chunk $P_{chunk_i}$, the input frame $I_{\text{input}}$, which is the final frame from the previous chunk or $I_0$ initially, is warped to the other poses of the chunk using the depth $D$ estimated by a pre-trained monocular model~\citep{bhat2023zoedepth}.

Within a chunk, the poses are indexed locally as $[R_j \,|\, t_j]$, $j=0,\dots,N-1$, and the first pose serves as the reference. With camera intrinsics $\mathbf{K}$, each pixel $p=(u,v)$ of $I_{\text{input}}$ is lifted to a 3D point in the world frame,
\begin{equation}
X(p) = R_0^{\top}\big(D(p)\,\mathbf{K}^{-1}\tilde{p} - t_0\big), \qquad \tilde{p}=(u,v,1)^{\top}.
\label{eq:unproject}
\end{equation}
The point is then projected into every other pose of the chunk,
\begin{equation}
\lambda_j\,\tilde{p}_j = \mathbf{K}\big(R_j X(p) + t_j\big), \qquad \tilde{p}_j=(p_j,1)^{\top}, \qquad j=1,\dots,N-1,
\label{eq:project}
\end{equation}
where $p_j$ is the pixel position of the point in camera $j$ and $\lambda_j$ its depth, following the standard pinhole model~\citep{zisserman2004multiple}. Only points with $\lambda_j>0$ and $p_j$ inside the image are kept. Since the poses are given by the camera path, Equations~\ref{eq:unproject} and~\ref{eq:project} need no per-frame alignment, which is what makes the warping registration-free.

Small gaps in the warped frames are first filled by the distance-transform step described below. Large holes and regions outside the source view are then filled with a pre-trained Stable Diffusion inpainting model~\citep{rombach2021high}. The resulting multi-view sequence serves as conditioning for the diffusion backbone.

Rebuilding the point cloud at each chunk boundary limits geometric distortion within each pass and reduces error accumulation, while the camera still moves continuously along the global trajectory. The number of passes follows from a bound on the camera rotation per pass. A path whose total rotation is $\Theta$ is split into $K=\lceil \Theta/\theta_{\max} \rceil$ passes of equal arc, with $\theta_{\max}=15^\circ$. A short path that rotates by less than $15^\circ$ is therefore generated in one pass, and a $60^\circ$ orbit in four. Section~\ref{sec:experiments} compares arcs from $10^\circ$ to $60^\circ$ on matched poses. Geometric error keeps falling as the arc shrinks, while run time grows with the number of passes, and $15^\circ$ gives a good balance between the two.

\label{sec:3.2} \vspace{1mm} \noindent \textbf{Efficient Registration-Free Warping.} In the original CamTrol pipeline~\citep{hou2024training}, each frame requires an iterative point cloud registration step to refine camera alignment. Although accurate, this loop is expensive, and with the interpolation that follows it the warping stage takes about $11.4$\,s per warped frame at $960\times704$ on an NVIDIA L40S GPU. CamTrol++ removes this per-frame optimization entirely. To keep warp quality without registration, warping is done in two stages. First, forward splatting~\citep{zwicker2002surface} spreads the color of each projected point over the $3\times3$ pixels around $p_j$,
\begin{equation}
\hat{I}_j(q) = \frac{\sum_{p} w(q,p_j)\, I_{\text{input}}(p)}{\sum_{p} w(q,p_j)}, \qquad w(q,p_j) = \frac{1}{1+|q_x-\lfloor p_j\rceil_x|+|q_y-\lfloor p_j\rceil_y|},
\label{eq:splat}
\end{equation}
where $\lfloor\cdot\rceil$ rounds to the nearest pixel and the sums run over points whose $3\times3$ footprint contains $q$. This reduces the small holes that arise when scattered points are interpolated onto the pixel grid, as in the interpolation-based warping of CamTrol. Second, pixels that receive no point take the color of the nearest filled pixel, found with a fast Euclidean distance transform~\citep{danielsson1980euclidean}, and pixels far from every projected point are marked as holes for inpainting.

This registration-free design produces clean, nearly hole-free warped frames in $1.9$\,s per frame. On the same GPU, resolution and frames, removing registration and replacing interpolation with splatting makes warping $6.1\times$ faster than in CamTrol~\citep{hou2024training} and a full pass including SVD refinement $3.2\times$ faster (Table~\ref{tab:supp_runtime}), at a quality comparable to linear interpolation (Table~\ref{tab:merged_color_interp}, both in Appendix~\ref{sec:supp_characterization}).

\label{sec:3.3} \vspace{1mm} \noindent \textbf{Perceptual Anchoring (Low-Frequency Appearance Control).} Autoregressive generation can introduce perceptual drift over time, particularly in color distribution (Figure~\ref{fig:figure_4_ablation}). As diffusion errors accumulate, frames can gradually deviate from the source image's appearance. A lightweight perceptual anchoring step, referred to as LAB anchoring in the rest of the paper, reduces this drift. Before the last frame of a chunk, $I_{\text{last}}$, is passed as input to the next chunk, its color statistics are matched to the original image $I_0$. Every pass boundary therefore restarts from the colors of the source image, which removes most of the drift across passes (chroma drift $2.49$ to $1.08$, Table~\ref{tab:merged_ablation_depth}). The same matching is also applied to the inpainted warped frames before they condition the backbone, so newly filled regions start with the colors of the source image. Matching at pass boundaries alone gives slightly lower geometric error at a small cost in color fidelity (Table~\ref{tab:merged_ablation_depth}).

As in color transfer~\citep{reinhard2002color}, which works in a decorrelated color space, both images are converted to the CIELAB color space, which separates lightness (L*) from chrominance (a*, b*). The L* channel of $I_{\text{last}}$ is kept, and histogram matching~\citep{chang1978optimal} is performed on the a* and b* channels with respect to $I_0$. The corrected frame is reconstructed by merging the preserved L* with the aligned chrominance channels and converting back to RGB. This step provides an additional constraint on low-frequency chromatic drift while preserving the geometric structure of the generated frame.

\label{sec:3.4} \vspace{1mm} \noindent \textbf{Geometry-Constrained Spatial Attention.} To improve geometric consistency during generation, epipolar-guided spatial attention~\citep{ye2025synthesizing} is added to the SVD decoder. Standard self-attention aggregates features globally, which can introduce inconsistent feature correspondences under viewpoint changes. Here attention is instead restricted to epipolar lines defined by the camera motion. For a query feature $\mathbf{q}$ at position $x$ in frame $j$, its epipolar line in the reference frame $0$ is $\ell = F_j\,\tilde{x}$, with the fundamental matrix
\begin{equation}
F_j = \mathbf{K}^{-\top}\,[t_{0j}]_{\times}\,R_{0j}\,\mathbf{K}^{-1}, \qquad R_{0j}=R_0R_j^{\top}, \qquad t_{0j}=t_0-R_{0j}\,t_j,
\label{eq:fundamental}
\end{equation}
where $[R_{0j}\,|\,t_{0j}]$ is the relative pose from camera $j$ to camera $0$, $[\cdot]_{\times}$ is the cross-product matrix and $\mathbf{K}$ is rescaled to the latent resolution. Rather than attending to all spatial positions, $\mathbf{q}$ attends only to key-value pairs $(k_s, v_s)$ sampled along this line. A total of $n_s = W/2$ points are sampled uniformly along the epipolar line, where $W$ is the latent width, and scaled dot-product attention is computed over these points. The geometry-constrained output is combined with standard spatial attention,\vspace{-0.7em}
\begin{equation}
    \label{eq:epipolar_blend}
    v_{out} = \alpha v_{epi} + (1-\alpha) v_{orig},
\end{equation}
where $v_{epi}$ denotes the epipolar-constrained output, $v_{orig}$ the standard attention output, and $\alpha=0.9$ balances geometric consistency and generative flexibility.

\begin{table*}[!t]
\centering
\captionof{table}{\textbf{Quantitative Comparison.} CamTrol++ is compared with training-free baselines on RealEstate10K~\citep{zhou2018stereo} and MegaScenes~\citep{tung2024megascenes} using $14$-frame and $56$-frame sequences. CamTrol++ provides the strongest overall temporal and perceptual consistency in the 56-frame setting, with substantial improvements in LPIPS, CLIPSIM, FPA, and Angular Consistency over WAVE. At $14$ frames, CamTrol++ is best on every RealEstate10K metric, while on MegaScenes it trails CamTrol on LPIPS and FPA, WAVE on Frobenius norm, and both baselines on rotation angle and angular consistency. On MegaScenes at $56$ frames, WAVE has lower rotation errors and ZMR. CamTrol generates one $14$-frame window, so it has no $56$-frame entry. Its longer sequences (Figures~\ref{fig:teaser}, \ref{fig:figure_3_qualitative_comp}, \ref{fig:supp_qual_megascene100} and Table~\ref{table:supp_table_5_28frames}) join independent $14$-frame segments. At $56$ frames, consecutive frames are close, so many pixels move less than the $0.75$ pixel ZMR threshold, and ZMR is not comparable with Tables~\ref{tab:merged_geometry_3dgs} and~\ref{tab:step_size_matched}. Like Table~\ref{tab:depth_ablation}, this table follows the WAVE protocol, so its values are not comparable with the matched orbit studies of Table~\ref{tab:merged_geometry_3dgs}, which adds NVS-Solver.}\vspace{-0.5em}
\resizebox{1.0\linewidth}{!}{
\def\arraystretch{1.2}
\begin{tabular}{llccccccccc}
\hline
\multicolumn{2}{c}{} &
  \multicolumn{2}{c}{Next} &
  \multicolumn{2}{c}{First} &
   &
   &
   &
   & \\ \cline{3-6}
\multicolumn{2}{c}{\multirow{-2}{*}{Method}} &
  LPIPS $\downarrow$ &
  CLIPSIM $\uparrow$ &
  LPIPS $\downarrow$ &
  CLIPSIM $\uparrow$ &
  \multirow{-2}{*}{\begin{tabular}[c]{@{}c@{}}Frob. Norm\\ (Rotation) $\downarrow$\end{tabular}} &
  \multirow{-2}{*}{\begin{tabular}[c]{@{}c@{}}Rot. Angle\\ Diff. $\downarrow$ \end{tabular}} &
  \multirow{-2}{*}{\begin{tabular}[c]{@{}c@{}}Angular \\ Cons. $\downarrow$\end{tabular}} &
  \multirow{-2}{*}{FPA $\uparrow$} &
  \multirow{-2}{*}{ZMR $\downarrow$} \\ \hline
\multicolumn{1}{l|}{} &
  \multicolumn{10}{l}{\textbf{\textit{RealEstate10K}}} \\
\multicolumn{1}{l|}{} &
  WAVE &
0.3214 &
0.9565 &
0.5515 &
0.9086 &
1.4870 &
1.1006 &
32.9670 &
0.456 &
0.240 \\
\multicolumn{1}{l|}{} &
  CamTrol &
0.2653   &
0.9607   &
0.5437   &
0.8460   &
2.1322   &
1.6381   &
28.5376   &
0.749  &
0.151  \\
\multicolumn{1}{l|}{} &
  \cellcolor[HTML]{EFEFEF}CamTrol++ &
  \cellcolor[HTML]{EFEFEF}\textbf{0.2644} &
  \cellcolor[HTML]{EFEFEF}\textbf{0.9729} &
  \cellcolor[HTML]{EFEFEF}\textbf{0.5218} &
  \cellcolor[HTML]{EFEFEF}\textbf{0.9193} &
  \cellcolor[HTML]{EFEFEF}\textbf{1.1438} &
  \cellcolor[HTML]{EFEFEF}\textbf{0.8812} &
  \cellcolor[HTML]{EFEFEF}\textbf{15.7735} &
  \cellcolor[HTML]{EFEFEF}\textbf{0.771} &
  \cellcolor[HTML]{EFEFEF}\textbf{0.149} \\ \cline{2-11}
\multicolumn{1}{l|}{} &
  \multicolumn{10}{l}{\textbf{\textit{MegaScenes}}} \\
\multicolumn{1}{l|}{} &
  WAVE & 
0.3460  &
0.9359  &
0.5399  &
0.8590  &
\textbf{1.5849}  &
\textbf{1.2128}  &
\textbf{22.0563}  &
0.490 &
0.265 \\
\multicolumn{1}{l|}{} &
  CamTrol &
\textbf{0.2197}  &
0.9648  &
\textbf{0.4382}  &
0.8972  &
1.8477  &
1.3999  &
22.7652  &
\textbf{0.764}  &
0.265 \\
\multicolumn{1}{l|}{\multirow{-8}{*}{\rotatebox[origin=c]{90}{14 Frames}}} &
  \cellcolor[HTML]{EFEFEF}CamTrol++ &
  \cellcolor[HTML]{EFEFEF}0.2219 &
  \cellcolor[HTML]{EFEFEF}\textbf{0.9688} &
  \cellcolor[HTML]{EFEFEF}0.4487 &
  \cellcolor[HTML]{EFEFEF}\textbf{0.9095} &
  \cellcolor[HTML]{EFEFEF}1.7564 &
  \cellcolor[HTML]{EFEFEF}1.4120 &
  \cellcolor[HTML]{EFEFEF}25.4760 &
  \cellcolor[HTML]{EFEFEF}0.757 &
  \cellcolor[HTML]{EFEFEF}\textbf{0.236} \\ \hline
\multicolumn{1}{l|}{} &
  \multicolumn{10}{l}{\textbf{\textit{RealEstate10K}}} \\
\multicolumn{1}{l|}{} &
  WAVE &
  0.2189 &
  0.9692 &
  0.5430 &
  0.9068 &
  1.5594 &
  1.1729 &
  6.9601 &
  0.331 &
  0.342 \\
\multicolumn{1}{l|}{} &
  \cellcolor[HTML]{EFEFEF}CamTrol++ &
  \cellcolor[HTML]{EFEFEF}\textbf{0.0888} &
  \cellcolor[HTML]{EFEFEF}\textbf{0.9895} &
  \cellcolor[HTML]{EFEFEF}\textbf{0.5069} &
  \cellcolor[HTML]{EFEFEF}\textbf{0.9174} &
  \cellcolor[HTML]{EFEFEF}\textbf{1.2796} &
  \cellcolor[HTML]{EFEFEF}\textbf{1.0085} &
  \cellcolor[HTML]{EFEFEF}\textbf{4.9173} &
  \cellcolor[HTML]{EFEFEF}\textbf{0.890} &
  \cellcolor[HTML]{EFEFEF}\textbf{0.315} \\ \cline{2-11}
\multicolumn{1}{l|}{} &
  \multicolumn{10}{l}{\textbf{\textit{MegaScenes}}} \\
\multicolumn{1}{l|}{} &
  WAVE &
 0.2408  &
 0.9539  &
 0.5335  &
 0.8600  &
 \textbf{1.7198}  &
 \textbf{1.3525}  &
 7.3332  &
 0.425 &
 \textbf{0.336} \\
\multicolumn{1}{l|}{\multirow{-6}{*}{\rotatebox[origin=c]{90}{56 Frames}}} &
  \cellcolor[HTML]{EFEFEF}CamTrol++ &
  \cellcolor[HTML]{EFEFEF}\textbf{0.0794} &
  \cellcolor[HTML]{EFEFEF}\textbf{0.9881} &
  \cellcolor[HTML]{EFEFEF}\textbf{0.4340} &
  \cellcolor[HTML]{EFEFEF}\textbf{0.9074} &
  \cellcolor[HTML]{EFEFEF}1.7682 &
  \cellcolor[HTML]{EFEFEF}1.5059 &
  \cellcolor[HTML]{EFEFEF}\textbf{4.2494} &
  \cellcolor[HTML]{EFEFEF}\textbf{0.893} &
  \cellcolor[HTML]{EFEFEF}0.401 \\ \hline
\end{tabular}}
\label{table:table_1}\vspace{-0.5em}
\end{table*}

The constraint is applied at the final spatial attention layer of the SVD decoder. Layer-wise analysis (Table~\ref{tab:supp_table_2}, Appendix~\ref{sec:supp_characterization}) shows that injection at layer $2$ suppresses motion, while layers $8$, $17$ and $23$ behave similarly, and layer $23$ gives the lowest ZMR among them. Although the constraint is applied within each chunk, it provides an additional geometric constraint during autoregressive generation. It acts as a light refinement on top of the bounded camera step and lowers epipolar error and LPIPS-\textit{next} at both step sizes of the component ablation (Table~\ref{tab:merged_ablation_depth}). Its gain is small, and in the full system it is offset by matching the inpainted frames (Section~\ref{sec:experiments}).
\section{Experiments}
\label{sec:experiments}

This section evaluates CamTrol++ on camera-controlled novel view synthesis. It first compares CamTrol++ with existing training-free methods on perceptual, temporal, motion and camera consistency, then on geometric consistency, downstream 3D reconstruction and fidelity to real camera motion. It then studies the components, the camera step size, a second video backbone, longer trajectories and depth errors. In each comparison, all methods share one camera path. The path is defined once, in world-to-camera form with the same intrinsics, and every method reads it from the same file.

\begin{table*}[!t]
\centering\caption{\textbf{Geometric Consistency and 3D Reconstruction on a $60^\circ$ Orbit.} CamTrol++ is compared with WAVE~\citep{park2025wave}, NVS-Solver~\citep{you2025nvssolver} and CamTrol~\citep{hou2024training}. All methods start from the same $20$ MegaScenes~\citep{tung2024megascenes} images, follow the same camera path with the same intrinsics, and are scored on the same $14$ poses. EAE and Sampson distance measure agreement with the commanded camera, FPA measures motion smoothness, and ZMR is the fraction of pixels that do not move between frames. Time is per scene on one L40S GPU. The right block fits 3D Gaussian Splatting~\citep{kerbl20233d} to each sequence with the commanded poses and reports PSNR, SSIM and LPIPS on four held-out views, together with the number of sequences that COLMAP registers in full. CamTrol++ gives the lowest geometric error and the best held-out reconstruction, and reaches the quality of CamTrol with the same pass structure in less than a third of the time.}\vspace{-0.5em}
{%
\resizebox{1.0\linewidth}{!}{
\def\arraystretch{1.3}
\begin{tabular}{lc|cccccc|cccc}
\toprule
 & & \multicolumn{6}{c|}{\textbf{Generated sequence}} & \multicolumn{4}{c}{\textbf{3D Gaussian Splatting, held-out views}} \\
\textbf{Method} & \textbf{Passes} & \textbf{LPIPS-\textit{next}} $\downarrow$ & \textbf{EAE} ($\times10^{-3}$) $\downarrow$ & \textbf{Sampson} ($\times10^{-4}$) $\downarrow$ & \textbf{FPA} $\uparrow$ & \textbf{ZMR} $\downarrow$ & \textbf{Time (s)} $\downarrow$ & \textbf{COLMAP} $\uparrow$ & \textbf{PSNR} $\uparrow$ & \textbf{SSIM} $\uparrow$ & \textbf{LPIPS} $\downarrow$ \\
\midrule
WAVE                 & 1 & 0.187 & 5.41 & 25478 & 0.577 & \textbf{0.047} & \textbf{24}  & 0/20  & 18.58 & 0.541 & 0.237 \\
NVS-Solver       & 1 & \textbf{0.109} & 4.27 & 15714 & 0.769 & 0.199 & 595 & 14/20 & 17.85 & 0.551 & 0.223 \\
CamTrol           & 1 & 0.140 & 5.80 & 27518 & 0.714 & 0.243 & 218 & 7/20  & 16.31 & 0.413 & 0.343 \\
CamTrol, $15^\circ$ passes & 4 & 0.134 & 2.94 & 5374 & \textbf{0.833} & 0.049 & 760 & \textbf{20/20} & 18.41 & 0.600 & 0.211 \\
CamTrol++, single pass                   & 1 & 0.137 & 6.56 & 44472 & 0.715 & 0.234 & 56  & 8/20  & 16.02 & 0.393 & 0.355 \\
\rowcolor[HTML]{EFEFEF}
CamTrol++                                & 4 & 0.124 & \textbf{2.86} & \textbf{4869} & 0.829 & 0.058 & 222 & \textbf{20/20} & \textbf{19.25} & \textbf{0.626} & \textbf{0.185} \\
\bottomrule
\end{tabular}}}
\label{tab:merged_geometry_3dgs}\vspace{-1.2em}
\end{table*}

\vspace{1mm} \noindent \textbf{Datasets and Baselines.} The evaluation uses two large-scale in-the-wild datasets, RealEstate10K~\citep{zhou2018stereo} and MegaScenes~\citep{tung2024megascenes}. For the comparison without ground truth (Table~\ref{table:table_1}), $308$ images are sampled at random from RealEstate10K and $98$ from MegaScenes. The orbit comparison and the controlled studies use a fixed set of $20$ MegaScenes images and a $60^\circ$ orbit that is scored on $14$ evenly spaced poses. Fidelity to real camera motion is measured on $44$ RealEstate10K clips, with camera poses estimated by VGGT~\citep{wang2025vggt} and scaled to the ZoeDepth depth of the first frame. MipNeRF360~\citep{barron2022mip} is used for 3D reconstruction, with two views from each of seven of its scenes. CamTrol++ is compared with three training-free baselines. (1) CamTrol~\citep{hou2024training}, which provides training-free camera control through an SVD-based warp-and-inpaint pipeline and is run both in its native single-window setting and with the same $15^\circ$ passes as CamTrol++, (2) WAVE~\citep{park2025wave}, an image-centric novel view synthesis method, run from its official code and MegaScenes checkpoint, and (3) NVS-Solver~\citep{you2025nvssolver}, which guides a frozen SVD-XT model with warped views at every sampling step, run from its official code. All methods receive the same input image, depth, intrinsics and camera path.

\vspace{1mm} \noindent \textbf{Metrics.} Following WAVE~\citep{park2025wave}, several aspects of the generated sequences are measured. (1) \textit{Temporal Consistency.} LPIPS-\textit{next}~\citep{zhang2018unreasonable,park2025wave} and CLIPSIM-\textit{next}~\citep{radford2021learning,park2025wave} measure perceptual and semantic similarity between consecutive frames ($t$ and $t-1$). (2) \textit{Drift from Source.} LPIPS-\textit{first} and CLIPSIM-\textit{first} compare each generated frame $I_t$ with the input image $I_0$. (3) \textit{Camera Accuracy (SfM).} Camera parameters are estimated with COLMAP~\citep{schoenberger2016sfm,schoenberger2016mvs}, and the evaluation reports Frobenius Norm, Rotation Angle Difference, and Angular Consistency following the WAVE evaluation protocol~\citep{park2025wave}. (4) \textit{Geometric Consistency.} Epipolar Alignment Error (EAE) and Sampson Distance~\citep{zisserman2004multiple} are computed from point matches between frames, with the fundamental matrix built from the camera poses. Low values mean that the generated frames agree with the camera path that was asked for. (5) \textit{Motion Fidelity.} Flow Persistence Agreement (FPA) and Zero-Motion Ratio (ZMR) are computed from dense RAFT optical flow~\citep{teed2020raft}. FPA averages the agreement in direction and magnitude between the flow of one frame pair and the next, so it reflects motion smoothness. ZMR is the fraction of pixels whose flow between consecutive frames is shorter than $0.75$ pixels at $256\times192$. A high ZMR means that large parts of the image do not move, which indicates a stalled camera. Since a static video also gives low LPIPS-\textit{next} and low epipolar error, these metrics are always read together with ZMR. (6) \textit{Fidelity and Reconstruction.} On RealEstate10K, PSNR, SSIM and LPIPS compare generated frames with the real frames at the same poses. For 3D reconstruction, 3D Gaussian Splatting models are trained with gsplat~\citep{ye2025gsplat} on the generated views, views $2$, $5$, $8$ and $11$ are held out for testing, and the number of sequences that COLMAP registers in full is reported. Exact definitions are given in Appendix~\ref{sec:supp_metrics}.

\vspace{1mm} \noindent \textbf{Implementation Details.} CamTrol++ runs SVD at $960\times704$ with $25$ denoising steps and at most $15^\circ$ of camera rotation per pass. Every method's outputs are resized to $256\times192$ for scoring, following WAVE~\citep{park2025wave}, and the camera intrinsics are rescaled to the same size. Run times are measured on one NVIDIA L40S GPU.


\vspace{1mm} \noindent \textbf{Quantitative Comparison.} CamTrol++ is first compared with existing training-free methods under the same camera trajectories, following the evaluation protocol of WAVE~\citep{park2025wave}. This is the largest comparison in the paper ($406$ images) and the largest one that measures camera accuracy with COLMAP, so it tests temporal consistency and camera drift at scale, while the geometric analysis below uses the matched orbit. Table~\ref{table:table_1} summarizes results on RealEstate10K and MegaScenes for both $14$-frame and $56$-frame sequences. For the $14$-frame setting, CamTrol generates $14$ frames directly, while WAVE and CamTrol++ generate $56$ frames and are uniformly subsampled to $14$ frames for evaluation. The main gains appear in the $56$-frame setting. On RealEstate10K, CamTrol++ reduces LPIPS-\textit{next} from $0.2189$ for WAVE to $0.0888$ and increases CLIPSIM-\textit{next} from $0.9692$ to $0.9895$. On MegaScenes, LPIPS-\textit{next} decreases from $0.2408$ to $0.0794$, while CLIPSIM-\textit{next} increases from $0.9539$ to $0.9881$. FPA rises from $0.425$ to $0.893$ on MegaScenes, though ZMR is higher there ($0.401$ against $0.336$). Angular Consistency improves from $6.96^\circ$ to $4.92^\circ$ on RealEstate10K, indicating reduced long-term camera drift. The gaps are larger at $56$ frames, but the $14$-frame rows of WAVE and CamTrol++ subsample the same videos, so this reflects closer frame spacing, not a longer horizon.


\vspace{1mm} \noindent \textbf{Geometric Consistency and 3D Reconstruction.} Table~\ref{tab:merged_geometry_3dgs} compares all methods on a $60^\circ$ orbit around $20$ MegaScenes scenes, scored on the same $14$ poses. CamTrol++ achieves the lowest EAE ($2.86\times10^{-3}$) and Sampson distance. Its EAE is $33$ to $56\%$ lower than that of every method that covers the orbit in one pass. NVS-Solver reaches EAE $4.27\times10^{-3}$, WAVE $5.41\times10^{-3}$ and CamTrol $5.80\times10^{-3}$. The lower LPIPS-\textit{next} of NVS-Solver comes with a ZMR of $0.199$, so a fifth of its pixels stay still between frames, against $0.058$ for CamTrol++. WAVE keeps the image moving but has the lowest FPA ($0.577$), which reflects flicker from its image-based generation.

The comparison with CamTrol isolates the effect of the bounded step. Running CamTrol with the same $15^\circ$ passes lowers its EAE from $5.80$ to $2.94\times10^{-3}$, close to CamTrol++. The bounded step is therefore the main source of the gain, and it carries over to another pipeline. CamTrol++ reaches this quality in $222$\,s per scene against $760$\,s for CamTrol with the same passes, at least a $3.2\times$ speedup from the registration-free warping.

\begin{table}[!t]
\centering
\caption{\textbf{Fidelity to Real Camera Motion.} $44$ RealEstate10K~\citep{zhou2018stereo} clips. Each method receives the first frame and the camera poses of the clip, and its frames are compared with the real frames at six target poses. The median clip rotates the camera by only $10^{\circ}$. CamTrol++ uses at most $15^{\circ}$ of rotation per pass, which gives one pass for $29$ clips, two passes for $14$ clips and four passes for the one clip that rotates by $50^{\circ}$. The number of passes is rounded up to $1$, $2$, $3$ or $6$ so that pass boundaries fall on the six target poses. CamTrol++ matches CamTrol at about half the run time, while NVS-Solver reaches the highest fidelity at eight times the run time. Using more passes than the motion requires lowers fidelity, as the fixed row of six $14$-frame passes shows.}\vspace{-0.5em}
{%
\resizebox{0.82\linewidth}{!}{
\def\arraystretch{1.1}
\begin{tabular}{lccccc}
\toprule
\textbf{Method} & \textbf{Passes} & \textbf{PSNR} $\uparrow$ & \textbf{SSIM} $\uparrow$ & \textbf{LPIPS} $\downarrow$ & \textbf{Time (s)} $\downarrow$ \\
\midrule
Input frame repeated        & n/a & 13.60 & 0.352 & 0.357 & n/a \\
WAVE             & 1   & 15.76 & 0.473 & 0.254 & \textbf{11} \\
CamTrol       & 1   & 17.63 & 0.576 & 0.179 & 141 \\
NVS-Solver   & 1   & \textbf{18.16} & \textbf{0.624} & \textbf{0.164} & 595 \\
CamTrol++, fixed $6$ passes          & 6   & 16.36 & 0.494 & 0.258 & 336 \\
\rowcolor[HTML]{EFEFEF}
CamTrol++                            & 1 to 6 & 17.64 & 0.568 & 0.179 & 74 \\
\bottomrule
\end{tabular}}}
\label{tab:re10k_fidelity}\vspace{-0.5em}
\end{table}

The right block of Table~\ref{tab:merged_geometry_3dgs} reports 3D reconstruction from the generated views. CamTrol++ gives the best held-out quality ($19.25$\,dB PSNR, $0.626$ SSIM, $0.185$ LPIPS), and COLMAP registers all $20$ of its sequences. Splitting the path improves reconstruction for both pipelines. Against a single pass, $15^\circ$ steps add $3.2$\,dB for CamTrol++ (better in $19$ of $20$ scenes) and $2.1$\,dB for CamTrol, and raise COLMAP registration from $8$ and $7$ of $20$ sequences to $20$ of $20$. WAVE sequences are not registered by COLMAP in any scene, which agrees with its high frame-to-frame inconsistency. On MipNeRF360, $15^\circ$ steps raise held-out PSNR from $16.98$ to $17.42$\,dB and SSIM from $0.317$ to $0.444$, lower LPIPS from $0.423$ to $0.296$ compared with a single pass, and let COLMAP register $12$ of $14$ sequences against $3$ of $14$ (Table~\ref{tab:supp_mipnerf360}, Appendix~\ref{sec:supp_characterization}).

\begin{table*}[!t]
\centering
\caption{\textbf{Component Ablation.} Mean over the same $20$ MegaScenes images and the same $14$ poses of a $60^\circ$ orbit. Components are added on top of small-step reprojection at two step sizes. The last two rows of each block are two placements of LAB anchoring, each added to the epipolar attention row. The shaded row is the configuration used for CamTrol++ in all other experiments. Epipolar attention lowers EAE, Sampson distance and LPIPS-\textit{next} at both step sizes. LAB anchoring at pass boundaries removes most of the color drift and saturation change, with chroma drift falling from $2.49$ to $1.08$ at $15^\circ$, and leaves geometry unchanged. Matching the inpainted frames as well gives the lowest chroma drift and saturation deviation at both step sizes, at a small cost in EAE and Sampson distance. Anchoring at pass boundaries alone is therefore a lighter variant with slightly lower geometric error. The step size itself has a larger effect on geometry than any component (EAE $2.87$ at $15^\circ$ against $3.74$ at $30^\circ$).}\vspace{-0.5em}
{%
\resizebox{\linewidth}{!}{
\def\arraystretch{1.12}
\begin{tabular}{lccccccc}
\toprule
\textbf{Configuration} & \begin{tabular}[c]{@{}c@{}}\textbf{LPIPS-\textit{next}}\\$\downarrow$\end{tabular} & \begin{tabular}[c]{@{}c@{}}\textbf{LPIPS-\textit{first}}\\$\downarrow$\end{tabular} & \begin{tabular}[c]{@{}c@{}}\textbf{EAE}\\($\times10^{-3}$) $\downarrow$\end{tabular} & \begin{tabular}[c]{@{}c@{}}\textbf{Sampson}\\($\times10^{-4}$) $\downarrow$\end{tabular} & \begin{tabular}[c]{@{}c@{}}\textbf{FPA}\\$\uparrow$\end{tabular} & \begin{tabular}[c]{@{}c@{}}\textbf{Chroma}\\\textbf{drift} $\downarrow$\end{tabular} & \begin{tabular}[c]{@{}c@{}}\textbf{Saturation}\\\textbf{dev.} $\downarrow$\end{tabular} \\
\midrule
\multicolumn{8}{l}{\textit{$15^\circ$ per pass (4 passes)}} \\
Small-step reprojection                      & 0.125 & 0.342 & 2.87 & 5107 & 0.826 & 2.63 & 0.519 \\
\quad + Epipolar attention                   & 0.122 & 0.337 & 2.81 & 4678 & \textbf{0.838} & 2.49 & 0.498 \\
\quad + LAB at boundaries only & \textbf{0.122} & \textbf{0.334} & \textbf{2.80} & \textbf{4537} & \textbf{0.838} & 1.08 & 0.155 \\
\rowcolor[HTML]{EFEFEF}
\quad + LAB at boundaries and inpainted frames (CamTrol++) & 0.124 & 0.336 & 2.86 & 4869 & 0.829 & \textbf{1.04} & \textbf{0.133} \\
\midrule
\multicolumn{8}{l}{\textit{$30^\circ$ per pass (2 passes)}} \\
Small-step reprojection                      & 0.130 & 0.323 & 3.74 & 8672 & \textbf{0.830} & 1.46 & 0.186 \\
\quad + Epipolar attention                   & 0.129 & 0.322 & 3.68 & 8470 & 0.818 & 1.45 & 0.179 \\
\quad + LAB at boundaries only & 0.129 & \textbf{0.321} & \textbf{3.65} & \textbf{8104} & 0.826 & 1.11 & 0.122 \\
\rowcolor[HTML]{EFEFEF}
\quad + LAB at boundaries and inpainted frames (CamTrol++) & \textbf{0.128} & 0.322 & 3.80 & 8836 & 0.819 & \textbf{1.04} & \textbf{0.113} \\
\bottomrule
\end{tabular}}}
\label{tab:merged_ablation_depth}
\vspace{-0.5em}
\end{table*}

\begin{figure}[t]
\centering
\includegraphics[width=\linewidth]{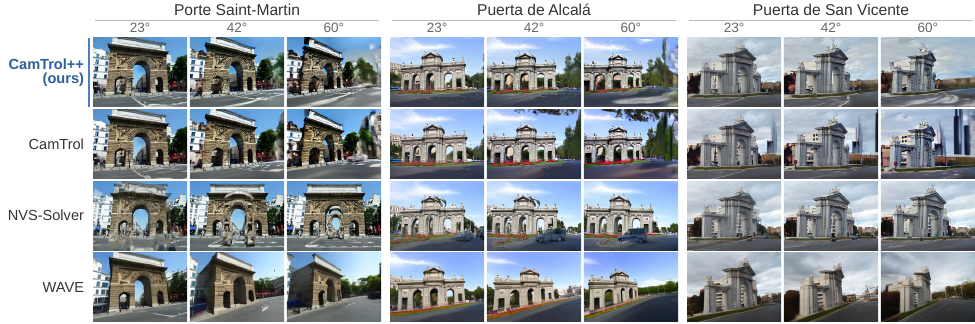}\vspace{-0.5em}
\caption{\textbf{Orbit on MegaScenes.} Three scenes rendered along the same $60^\circ$ orbit, shown at $23^\circ$, $42^\circ$ and $60^\circ$ from the input view. CamTrol++ turns around each monument and keeps its arches, reliefs and statues sharp and in place. CamTrol with the same $15^\circ$ passes adds structures that are not in the scene, such as the tall towers beside Puerta de San Vicente and the tree branches above Puerta de Alcal\'a. NVS-Solver barely changes the viewpoint and inserts objects, such as a second arch inside Porte Saint-Martin and a van in front of Puerta de Alcal\'a. WAVE moves the camera but redraws the monuments, skewing Porte Saint-Martin and flattening the gate of Puerta de San Vicente.}\vspace{-1.5em}
\label{fig:qual_orbit}
\end{figure}

\begin{figure}[t]
\centering
\includegraphics[width=\linewidth]{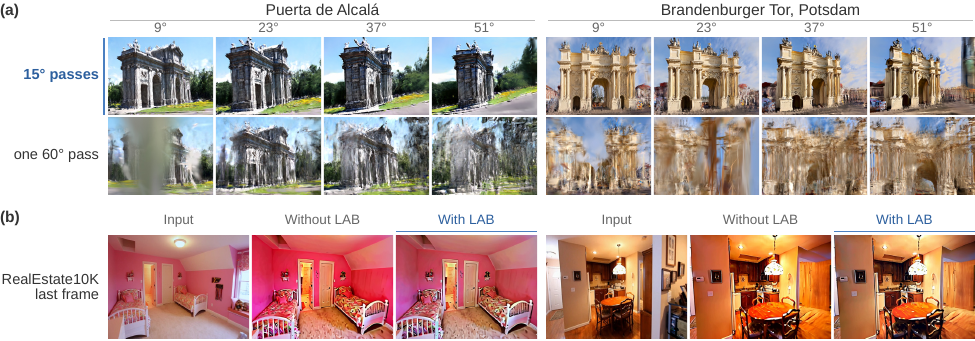}\vspace{-0.5em}
\caption{\textbf{Downstream 3D Reconstruction and Color Stability.} (a) Held-out views of a 3D Gaussian Splatting model fitted to CamTrol++ frames at the camera poses. Frames from $15^\circ$ passes give a clean, consistent reconstruction, while frames from one $60^\circ$ pass disagree with each other and the fitted model breaks into streaks and floaters. (b) Last generated frame of two RealEstate10K clips without and with LAB color matching at the pass boundaries. Without matching, the colors drift towards stronger magenta and orange tones, and with matching they stay closer to the input frame.}\vspace{-0.8em}
\label{fig:qual_3dgs}
\end{figure}

\begin{figure}[!t]
\centering
\begin{subfigure}[t]{0.39\linewidth}
  \vspace{0pt}\centering
  \includegraphics[width=\linewidth]{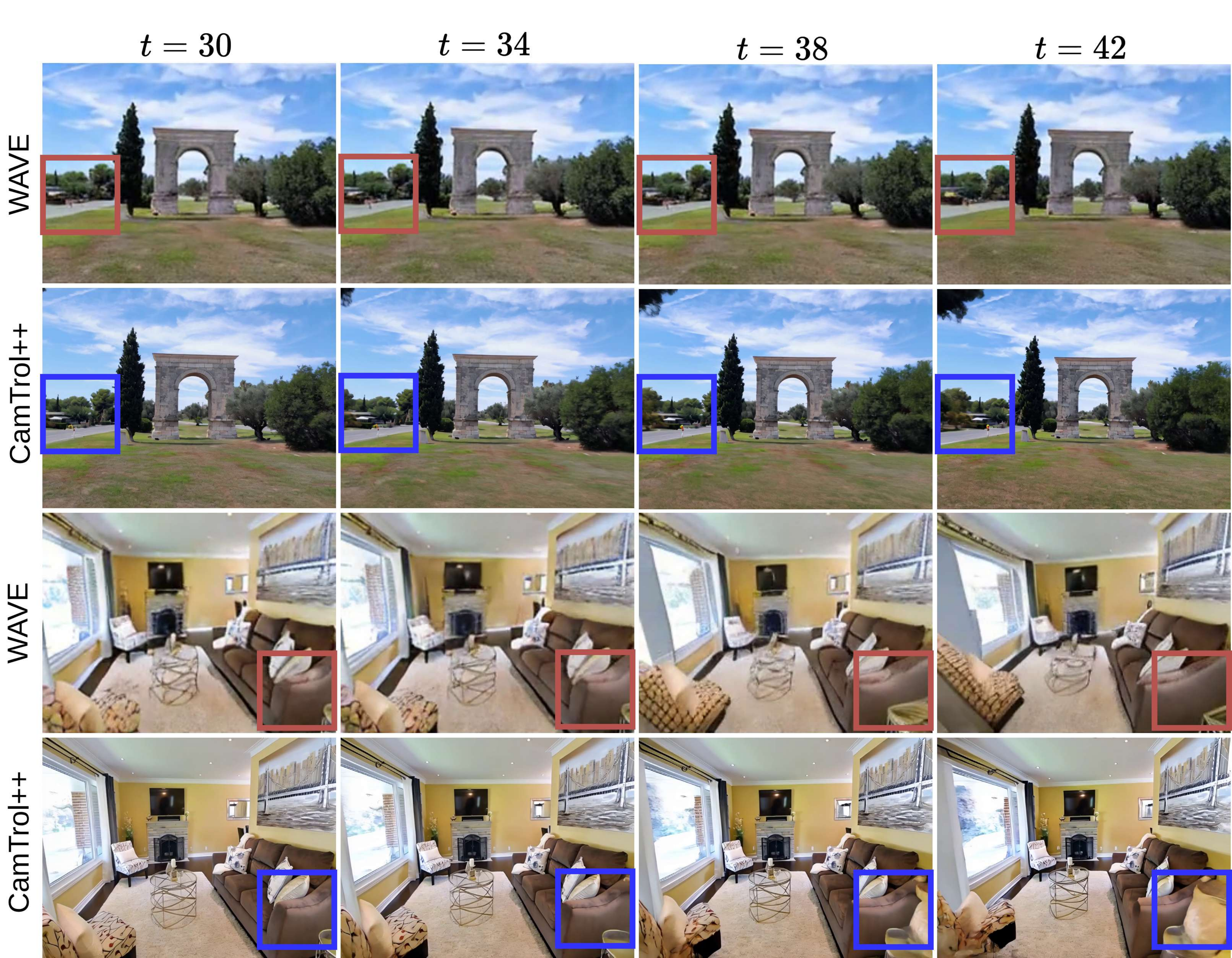}
  \caption{Static frames}\vspace{-0.5em}
  \label{fig:figure_6_static_frames}
\end{subfigure}\hfill
\begin{subfigure}[t]{0.58\linewidth}
  \vspace{0pt}\centering
  \includegraphics[width=\linewidth]{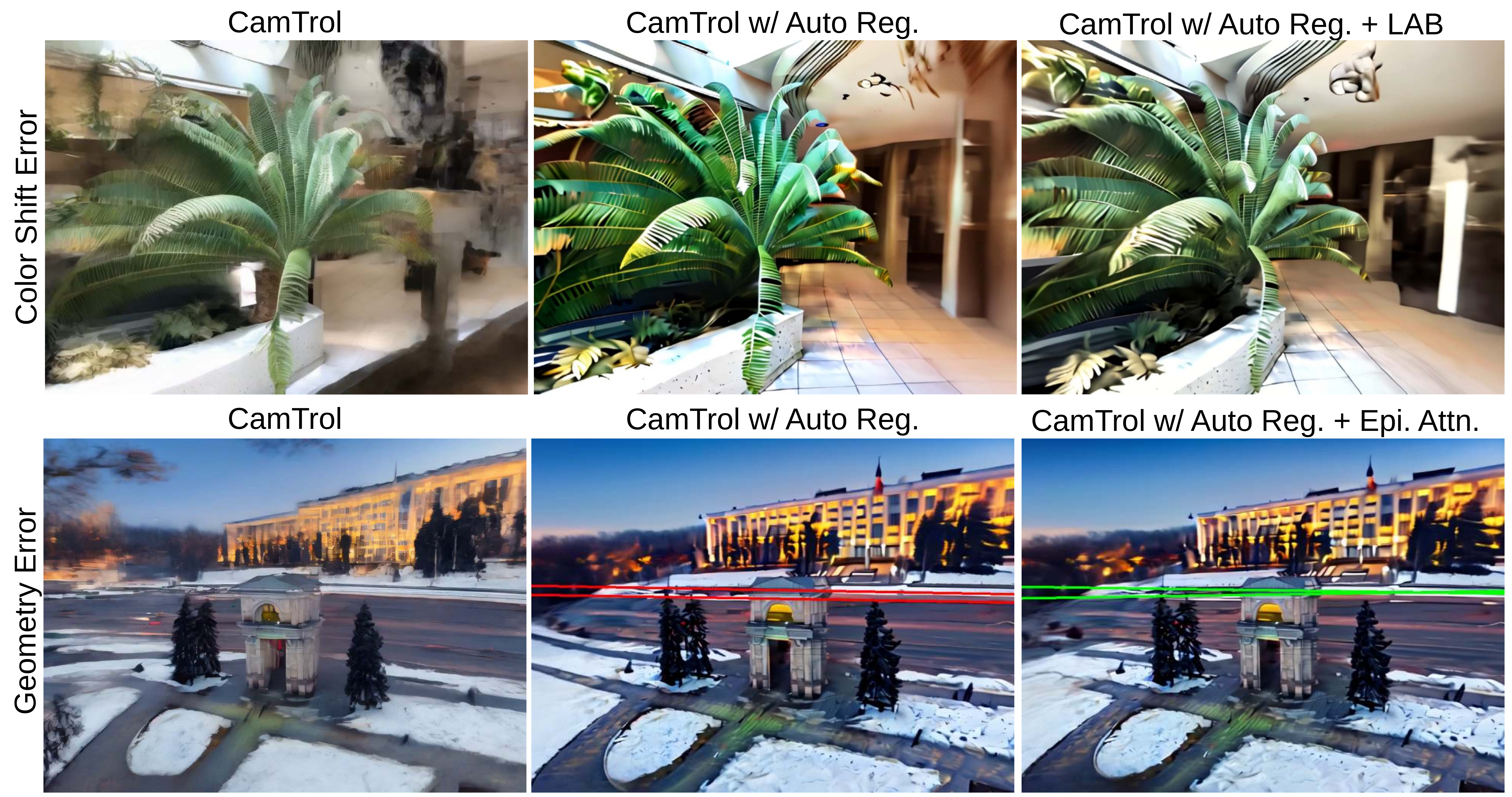}
  \caption{Visual ablation}\vspace{-0.5em}
  \label{fig:figure_4_ablation}
\end{subfigure}
\vspace{-0.5em}
\caption{\textbf{Static Frames and Visual Ablation.} \textit{(a) Analysis of Static Frames.} This figure illustrates a common failure mode of image-centric baselines such as WAVE~\citep{park2025wave} in a $56$-frame sequence. Frames from $t=30$ to $t=42$ are shown at a $4$-frame interval. In the WAVE results, the boxed regions (red) remain static, while CamTrol++ (blue boxes) follows the change in viewpoint. WAVE's ZMR is higher in three of the four settings of Table~\ref{table:table_1}; on MegaScenes at $56$ frames and in Table~\ref{tab:merged_geometry_3dgs} it is lower, and its low FPA ($0.425$ against $0.893$ on MegaScenes) indicates inconsistent motion. \textit{(b) Visual Ablation Study.} The effects of LAB anchoring and epipolar attention are shown. \textit{Top row, color shift error.} The CamTrol baseline~\citep{hou2024training} (left) shows noticeable color distortion, while the autoregressive setup (+~Auto Reg.) reduces this effect but still shows some drift. Adding LAB anchoring (right) further stabilizes the color appearance. \textit{Bottom row, geometry error.} The autoregressive setup (center) shows perspective deviations, visible from the diverging red lines. With epipolar attention (right), the lines remain consistent with the expected geometry, with the parallel lines (green) converging toward a stable vanishing point.}
\label{fig:pair_static_ablation}\vspace{-0.5em}
\end{figure}

\vspace{1mm} \noindent \textbf{Fidelity to Real Camera Motion.} The metrics above measure agreement with the commanded path. Table~\ref{tab:re10k_fidelity} compares the generated frames with real frames. On $44$ RealEstate10K clips, each method receives the first frame and the VGGT poses of the clip, and its frames are compared with the real frames at six target poses. The median clip rotates the camera by only $10^{\circ}$, so the $15^{\circ}$ bound gives a single pass for $29$ of the $44$ clips, two for $14$ clips and four for the one clip that rotates by $50^{\circ}$ (six after rounding, Table~\ref{tab:re10k_fidelity}). CamTrol++ reaches $17.64$\,dB PSNR and $0.179$ LPIPS, on par with CamTrol, at $74$\,s per clip against $141$\,s. NVS-Solver reaches the highest fidelity on these short paths ($18.16$\,dB) through gradient guidance at every sampling step, at $595$\,s per clip. Every method is far closer to the real frames than a static copy of the input ($13.60$\,dB). The fixed six-pass row always uses six passes, whatever the rotation of the clip, and shows why the number of passes follows the rotation of the path. Splitting a short path into six passes of $14$ frames, more than the bound requires, lowers PSNR by $1.3$\,dB, since each extra pass regenerates from an already generated frame.

\begin{figure}[t]
\centering
\includegraphics[width=\linewidth]{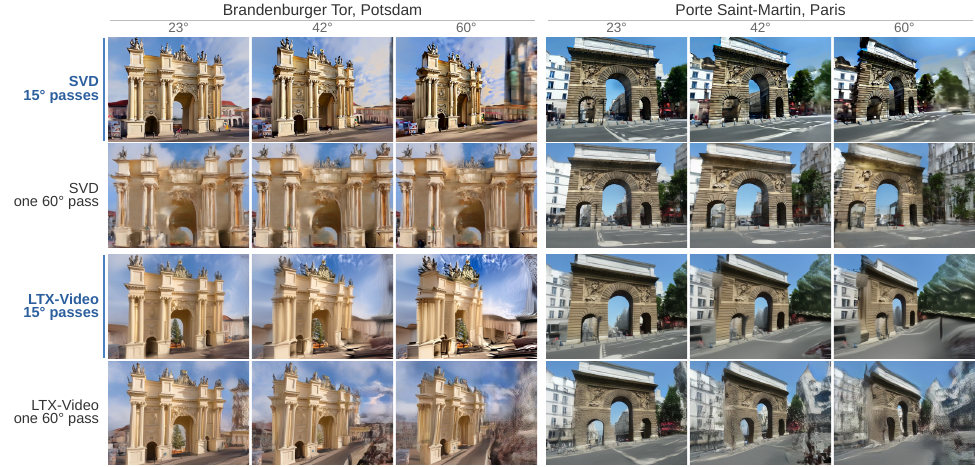}\vspace{-0.5em}
\caption{\textbf{Pass Size on Two Backbones.} The same $60^\circ$ orbit is generated either as four $15^\circ$ passes or as one $60^\circ$ pass, with SVD and with LTX-Video. The LTX-Video rows use reprojection alone, without attention control or color matching. With $15^\circ$ passes both backbones turn around the monument and keep its overall shape. With one large pass, SVD freezes, so the camera barely moves and the facade melts into a flat texture, while LTX-Video fills new regions with noisy, rock-like structures.}\vspace{-1.0em}
\label{fig:qual_stepsize}
\end{figure}

\vspace{1mm} \noindent \textbf{Qualitative Analysis.} Figure~\ref{fig:teaser} illustrates challenging $60^{\circ}$ rotations. CamTrol++ maintains sharp structure and coherent motion, while CamTrol exhibits blur and color drift, and WAVE produces geometric distortions and flicker. Figure~\ref{fig:figure_3_qualitative_comp} further shows $56$-frame sequences on MegaScenes and RealEstate10K, with more in Appendix~\ref{sec:add_results}. CamTrol++ keeps a consistent perspective and appearance and follows the camera pose, while WAVE can produce static or sliding backgrounds that do not follow it (Figure~\ref{fig:figure_6_static_frames}). Figure~\ref{fig:qual_orbit} compares all methods at the same poses of the $60^\circ$ orbit. CamTrol++ keeps the shape of each monument as the camera turns, CamTrol adds structures that are not in the scene, NVS-Solver inserts duplicated arches and vehicles or barely changes the viewpoint, and WAVE skews and flattens the gates. Appendix~\ref{sec:supp_characterization} shows the same comparison against real RealEstate10K frames (Figure~\ref{fig:qual_re10k}). Figure~\ref{fig:qual_stepsize} shows why the arc per pass matters for both backbones. With one $60^\circ$ pass SVD freezes the camera and LTX-Video fills new regions with noisy structures, while $15^\circ$ passes turn around the monument. The same difference carries over to 3D reconstruction in Figure~\ref{fig:qual_3dgs}, where Gaussian Splatting fitted to $15^\circ$ passes renders clean held-out views and the single pass breaks into streaks and floaters. The same figure shows the color drift that anchoring at pass boundaries largely removes.


\begin{figure}[!t]
\centering
\includegraphics[width=\linewidth]{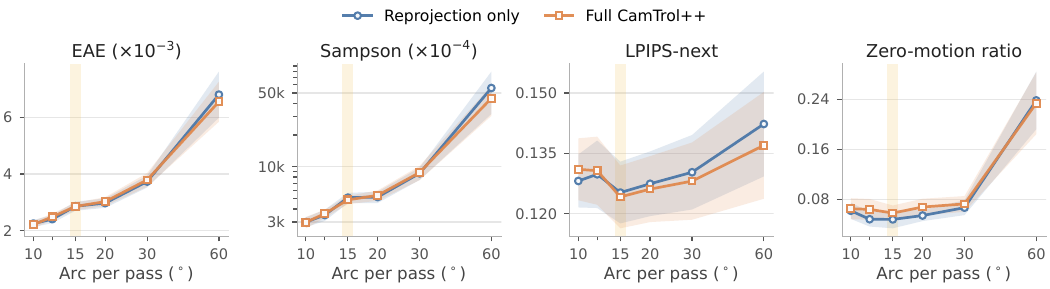}\vspace{-1.0em}
\vspace{-0.5em}
\captionof{figure}{\textbf{Effect of Camera Step Size.} Every arc size is generated along the same $60^\circ$ orbit and scored on the same $14$ poses ($n=20$ MegaScenes images). Lines show the mean and shaded bands show $\pm1$ SEM. Blue uses reprojection only, with epipolar attention and LAB anchoring turned off. Orange is the full CamTrol++. Geometric error grows with the arc. The single $60^\circ$ pass is also the only setting where large parts of the image stop moving (high ZMR). LPIPS-\textit{next} stays nearly flat and is therefore not used to choose the step size. The yellow band marks the $15^\circ$ default.}
\label{fig:arc_size_validation}
\vspace{-0.5em}
\end{figure}

\vspace{1mm} \noindent
\textbf{Ablation Study.} Table~\ref{tab:merged_ablation_depth} adds the components one at a time on top of small-step reprojection, at $15^\circ$ and $30^\circ$ per pass, on the same scenes and poses. Epipolar attention lowers EAE, Sampson distance and LPIPS-\textit{next} at both step sizes, for example Sampson distance from $5107$ to $4678\times10^{-4}$ at $15^\circ$. LAB anchoring at pass boundaries targets appearance. It lowers chroma drift from $2.49$ to $1.08$ and saturation deviation from $0.498$ to $0.155$ at $15^\circ$, with saturation closer to the source in all $20$ scenes, while geometry stays unchanged. Its effect is larger at $15^\circ$ than at $30^\circ$, since four passes have more boundaries at which drift can build up. Matching the inpainted frames as well gives the lowest chroma drift and saturation deviation at both step sizes ($1.04$ and $0.133$ at $15^\circ$), at a small cost in geometry (Sampson distance $4869$ against $4537\times10^{-4}$ and EAE $2.86$ against $2.80$). CamTrol++ uses both placements, so that newly inpainted regions also start from the colors of the source image, and all other results in this paper are reported with this configuration. Anchoring at pass boundaries alone remains a lighter variant for applications that favor geometric accuracy over color fidelity. In the full system, the small geometric gain of epipolar attention is offset by matching the inpainted frames, so the full system and reprojection only reach about the same geometric error at every arc (Table~\ref{tab:step_size_matched}). The components therefore mainly improve color, and the bound on the arc per pass carries the geometric gain. Boundary anchoring also holds against real frames. On the $44$ RealEstate10K clips, it lowers the color error to the real targets from $9.53$ to $8.97$ and improves $39$ of $44$ clips. A simple mean and variance transfer~\citep{reinhard2002color} at the same place performs on par ($8.94$, Table~\ref{tab:supp_color_placement}), which shows that the placement at pass boundaries is what matters. Figure~\ref{fig:figure_4_ablation} shows qualitative examples. Without LAB anchoring, the color tone gradually shifts across the sequence, while removing epipolar attention can lead to less consistent perspective structure.


\begin{table}[!t]
\centering
\caption{\textbf{Matched Camera Step Size.} Mean $\pm$ SEM over the same $20$ MegaScenes images for every row. All settings follow one $60^\circ$ orbit and are scored on the same $14$ poses, so only the arc per pass changes. EAE and Sampson distance grow steadily with the arc in both blocks. The single pass also leaves much more of the image static (ZMR $0.234$ to $0.238$ against at most $0.073$). LPIPS-\textit{next} stays within $0.124$ to $0.142$ for every setting. Time is the mean generation time per scene on one L40S GPU.}\vspace{-0.5em}
{%
\resizebox{\linewidth}{!}{
\def\arraystretch{1.1}
\begin{tabular}{lccccccc}
\toprule
\textbf{Arc per pass (deg)} & \textbf{Passes} & \textbf{LPIPS-\textit{next}} $\downarrow$ & \textbf{EAE} ($\times10^{-3}$) $\downarrow$ & \textbf{Sampson} ($\times10^{-4}$) $\downarrow$ & \textbf{FPA} $\uparrow$ & \textbf{ZMR} $\downarrow$ & \textbf{Time (s)} \\
\midrule
\multicolumn{8}{l}{\textit{Reprojection only}} \\
10 & 6 & $0.128 \pm 0.007$ & $2.26 \pm 0.12$ & $3000 \pm 321$ & $0.831 \pm 0.016$ & $0.062 \pm 0.013$ & 293 \\
12 & 5 & $0.130 \pm 0.008$ & $2.41 \pm 0.14$ & $3496 \pm 423$ & $0.835 \pm 0.015$ & $0.048 \pm 0.012$ & 244 \\
15 (default) & 4 & $0.125 \pm 0.008$ & $2.87 \pm 0.16$ & $5107 \pm 562$ & $0.826 \pm 0.019$ & $0.048 \pm 0.014$ & 195 \\
20 & 3 & $0.127 \pm 0.008$ & $2.98 \pm 0.17$ & $5174 \pm 631$ & $0.830 \pm 0.014$ & $0.054 \pm 0.009$ & 146 \\
30 & 2 & $0.130 \pm 0.009$ & $3.74 \pm 0.21$ & $8672 \pm 1160$ & $0.830 \pm 0.013$ & $0.067 \pm 0.012$ & 97 \\
60 (single pass) & 1 & $0.142 \pm 0.013$ & $6.81 \pm 0.81$ & $55579 \pm 23972$ & $0.700 \pm 0.041$ & $0.238 \pm 0.046$ & 49 \\
\midrule
\multicolumn{8}{l}{\textit{Full CamTrol++}} \\
10 & 6 & $0.131 \pm 0.008$ & $2.22 \pm 0.13$ & $2988 \pm 368$ & $0.838 \pm 0.015$ & $0.065 \pm 0.017$ & 332 \\
12 & 5 & $0.131 \pm 0.008$ & $2.49 \pm 0.11$ & $3604 \pm 322$ & $0.820 \pm 0.015$ & $0.064 \pm 0.017$ & 277 \\
15 (default) & 4 & $0.124 \pm 0.008$ & $2.86 \pm 0.13$ & $4869 \pm 431$ & $0.829 \pm 0.014$ & $0.058 \pm 0.012$ & 222 \\
20 & 3 & $0.126 \pm 0.008$ & $3.03 \pm 0.16$ & $5368 \pm 600$ & $0.827 \pm 0.017$ & $0.068 \pm 0.013$ & 166 \\
30 & 2 & $0.128 \pm 0.010$ & $3.80 \pm 0.24$ & $8836 \pm 1354$ & $0.819 \pm 0.019$ & $0.073 \pm 0.013$ & 111 \\
60 (single pass) & 1 & $0.137 \pm 0.013$ & $6.56 \pm 0.71$ & $44472 \pm 14160$ & $0.715 \pm 0.039$ & $0.234 \pm 0.050$ & 56 \\
\bottomrule
\end{tabular}}}
\label{tab:step_size_matched}\vspace{-0.8em}
\end{table}

\begin{table}[!t]
\centering
\caption{\textbf{Second Backbone.} The small-step procedure with LTX-Video~\citep{hacohen2024ltx} as the refiner, on the same $20$ MegaScenes images and the same $14$ scored poses as Table~\ref{tab:step_size_matched}. 3D Gaussian Splatting is fitted for the default step and the single pass, with commanded poses and four held-out views. Multi-pass generation reaches lower EAE than the single pass, and $15^\circ$ passes reach a $1.7$\,dB higher held-out PSNR. The lowest EAE lies at $20^\circ$ to $30^\circ$, and arcs below $15^\circ$ leave more of the image static, so the best arc per pass depends on the backbone.}\vspace{-0.5em}
{%
\resizebox{\linewidth}{!}{
\def\arraystretch{1.1}
\begin{tabular}{lcccccccc}
\toprule
\textbf{Arc per pass (deg)} & \textbf{Passes} & \textbf{LPIPS-\textit{next}} $\downarrow$ & \textbf{EAE} ($\times10^{-3}$) $\downarrow$ & \textbf{FPA} $\uparrow$ & \textbf{ZMR} $\downarrow$ & \textbf{3DGS PSNR} $\uparrow$ & \textbf{COLMAP} $\uparrow$ & \textbf{Time (s)} \\
\midrule
10 & 6 & 0.099 & 3.18 & 0.487 & 0.341 & n/a & n/a & 299 \\
12 & 5 & 0.110 & 3.09 & 0.570 & 0.292 & n/a & n/a & 249 \\
\rowcolor[HTML]{EFEFEF}
15 (default) & 4 & 0.116 & 2.63 & 0.688 & 0.196 & \textbf{20.93} & \textbf{19/20} & 199 \\
20 & 3 & 0.127 & 2.44 & 0.769 & 0.140 & n/a & n/a & 149 \\
30 & 2 & 0.133 & \textbf{2.31} & 0.836 & 0.080 & n/a & n/a & 99 \\
60 (single pass) & 1 & 0.142 & 3.25 & \textbf{0.842} & \textbf{0.024} & 19.23 & 17/20 & 49 \\
\bottomrule
\end{tabular}}}
\label{tab:ltx}\vspace{-0.5em}
\end{table}

\vspace{1mm} \noindent \textbf{Camera Step Size.} The bound on the arc per pass is the main design choice in CamTrol++. A fair test of its effect holds the camera path and the scored poses fixed. Every setting therefore follows the same $60^\circ$ orbit, and only the arc handled by each pass changes, from $10^\circ$ (six passes) to $60^\circ$ (a single pass). All settings are scored on the same $14$ poses of the orbit, on the same $20$ MegaScenes images. Figure~\ref{fig:arc_size_validation} and Table~\ref{tab:step_size_matched} show the results, both with reprojection only and with the full system.

Geometric error rises steadily with the arc per pass. For the full system, EAE grows from $2.22\times10^{-3}$ at $10^\circ$ to $2.86\times10^{-3}$ at $15^\circ$ and $6.56\times10^{-3}$ for the single pass, and Sampson distance follows the same order. Every multi-pass setting is better than the single pass on both metrics, and $10^\circ$ is still better than $15^\circ$. The single pass also fails in a second way. In its frames, $23\%$ of pixels barely move, against $6$ to $7\%$ for the multi-pass settings, and its FPA drops from about $0.83$ to $0.72$. LPIPS-\textit{next} stays between $0.124$ and $0.142$ for every setting, so it does not reflect these differences. A single pass of the backbone is therefore both less consistent with the camera path and more likely to stall. The warp shows why. On the same images, a $15^\circ$ pass leaves $5.2\%$ of each warped frame to the inpainter on average, against $32.5\%$ for a single $60^\circ$ pass, and $64.6\%$ of its last frame (Figure~\ref{fig:supp_hole_fraction}, Appendix~\ref{sec:supp_characterization}). A large arc thus hands the backbone frames that are largely invented, which it cannot bring back to the camera path.

Smaller arcs keep improving geometry but cost more time. One scene takes $56$\,s as a single pass, $222$\,s at $15^\circ$ and $332$\,s at $10^\circ$. The $15^\circ$ bound is chosen as a balance between accuracy and run time.


\vspace{1mm} \noindent \textbf{Second Backbone.} The procedure does not depend on SVD. Table~\ref{tab:ltx} applies it to LTX-Video~\citep{hacohen2024ltx}, a diffusion transformer, run here with $25$ frames per pass, on the same orbit, scenes and scored poses. The refinement strength is chosen on three separate calibration scenes and then fixed. With LTX-Video, $15^\circ$ passes improve held-out 3D reconstruction over a single pass by $1.7$\,dB PSNR ($20.93$ against $19.23$) and lower EAE from $3.25$ to $2.63\times10^{-3}$, with the lowest EAE at $20^\circ$ to $30^\circ$. LTX-Video covers a $60^\circ$ orbit in a single pass with half the epipolar error of SVD in the same reprojection-only setting ($3.25$ against $6.81\times10^{-3}$), and very short arcs make it hold more of the image still (ZMR $0.341$ at $10^\circ$). Bounding the arc per pass therefore helps both backbones, and the best bound depends on the range of motion that the backbone handles in one pass.

\begin{figure}[!t]
\centering
\begin{subfigure}[t]{0.35\linewidth}
  \vspace{0pt}\centering
  \includegraphics[width=0.85\linewidth]{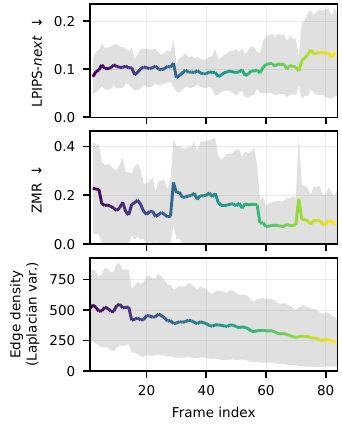}
  \caption{Long-horizon generation}
  \label{fig:operating_envelope}
\end{subfigure}\hfill
\begin{subfigure}[t]{0.61\linewidth}
  \vspace{0pt}\centering
  \includegraphics[width=0.85\linewidth]{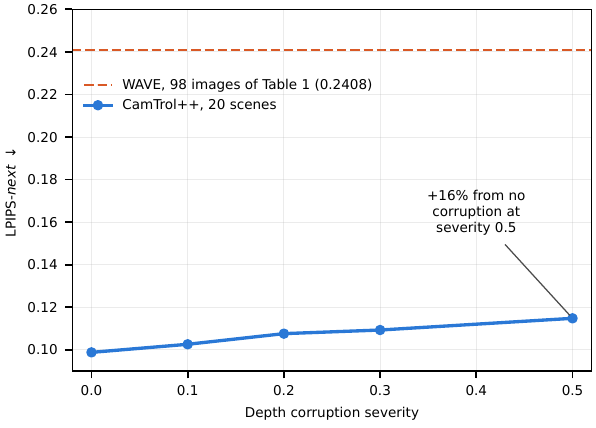}
  \caption{Depth robustness}
  \label{fig:depth_vs_wave}
\end{subfigure}
\vspace{-0.5em}
\caption{\textbf{Long-Horizon Generation and Depth Robustness.} \textit{(a) Long-Horizon Generation.} Mean $\pm$ std of three degradation signals (LPIPS-\textit{next}, ZMR, and edge density measured as the variance of the Laplacian of each frame, with bands clipped to the valid range) across an 84-frame single-direction extension, $n=20$ scenes from MegaScenes using the full configuration. Edge density declines steadily over the sequence, LPIPS-\textit{next} rises after about frame $70$, and ZMR does not increase. \textit{(b) Depth Robustness against a Training-Free Baseline.} LPIPS-\textit{next} under synthetic depth corruption applied to ZoeDepth's output, with corruption severity ranging from $0$ to $0.5$, compared against WAVE's 56-frame MegaScenes baseline ($0.2408$, dashed). Performance degrades gradually as the corruption becomes stronger. The sweep uses the $20$ scenes of the matched studies along the two-direction path of Section~\ref{sec:experiments} (uncorrupted LPIPS-\textit{next} $0.099$), while the WAVE line comes from the $98$ images of Table~\ref{table:table_1} and serves as a reference. At severity $0.5$ without missing pixels, LPIPS-\textit{next} rises by about $16\%$ to $0.115$.}
\label{fig:pair_longhorizon_depth}
\vspace{-1.0em}
\end{figure}

\vspace{1mm} \noindent \textbf{Long-Horizon Generation.} CamTrol++ is also run beyond the standard $56$-frame setting, with sequences of up to $84$ frames. Figure~\ref{fig:operating_envelope} shows the change in perceptual consistency, motion, and image detail as the sequence becomes longer. Image detail declines steadily over the sequence, LPIPS-\textit{next} rises only after about frame $70$, and ZMR does not increase. Beyond $56$ frames, a heuristic check of the $20$ extended sequences finds loss of texture and fine detail in $13$ scenes, appearance drift in $7$ and motion freezing in $5$, so detail loss is the most common failure.

\begin{table}[!t]
\centering
\caption{\textbf{Depth Estimation Ablation.} Comparison of different monocular depth predictors in the $56$-frame MegaScenes setting of Table~\ref{table:table_1}. ZoeDepth gives the best LPIPS-\textit{next}, CLIPSIM-\textit{next} and angular consistency, DepthPro gives the lowest rotation errors, and the gap to WAVE narrows with the relative depth of Marigold.}\vspace{-0.5em}
\resizebox{\linewidth}{!}{
\def\arraystretch{1.4}
\begin{tabular}{lccccc}
\toprule
Method & LPIPS-\textit{next} $\downarrow$ & CLIPSIM-\textit{next} $\uparrow$ &
Frob. Norm (Rot.) $\downarrow$ &
Rot. Angle Diff. $\downarrow$ &
Angular Cons. $\downarrow$ \\ \midrule
ZoeDepth~\citep{bhat2023zoedepth} & \textbf{0.0794} & \textbf{0.9881} &
1.7682 & 1.5059 & \textbf{4.2494} \\
DepthPro~\citep{bochkovskii2024depth} & 0.1036 & 0.9847 &
\textbf{1.7338} & \textbf{1.4255} & 6.8475 \\
DA3~\citep{lin2025depth} & 0.1387 & 0.9805 &
1.9491 & 1.4628 & 4.5052 \\
Marigold~\citep{ke2025marigold} & 0.2084 & 0.9619 &
2.4851 & 2.0219 & 6.2351 \\ \bottomrule
\end{tabular}
}\vspace{-0.3em}
\label{tab:depth_ablation}\vspace{-0.5em}
\end{table}
\begin{table}[!t]
\centering
\caption{\textbf{Structured Depth Failures.} Each failure is applied to the ZoeDepth map before warping, on the same $20$ MegaScenes images and $14$ scored poses. A depth scale that is too small is the failure that matters, since it enlarges the parallax of every warp. Small steps limit its effect to $+1.32$, against $+5.83$ for a single pass. Under every failure, $15^\circ$ steps keep a much lower EAE than the single pass. Fully flat depth keeps LPIPS-\textit{next} close to the default, and its ZMR stays far below the single pass (see text).}\vspace{-0.5em}
{%
\resizebox{0.8\linewidth}{!}{
\def\arraystretch{1.1}
\begin{tabular}{lcccc}
\toprule
 & \multicolumn{2}{c}{\textbf{Change in EAE} ($\times10^{-3}$)} & \multicolumn{2}{c}{\textbf{EAE under the failure} ($\times10^{-3}$)} \\
\cmidrule(lr){2-3}\cmidrule(lr){4-5}
\textbf{Depth failure} & $15^\circ$ & single pass & $15^\circ$ & single pass \\
\midrule
None (ZoeDepth)                    & 0.00  & 0.00  & 2.86 & 6.56 \\
Collapse halfway to a plane        & $-0.18$ & $-0.88$ & 2.68 & 5.68 \\
Collapse to a quarter of the range & $-0.06$ & $-1.06$ & 2.80 & 5.50 \\
Fully flat                         & $-0.17$ & $+1.35$ & 2.69 & 7.91 \\
Global scale $\times0.5$           & $+1.32$ & $+5.83$ & 4.18 & 12.39 \\
Global scale $\times2$             & $-0.62$ & $-2.73$ & 2.24 & 3.83 \\
Edge blur, $8$\,px                 & $-0.13$ & $-0.41$ & 2.73 & 6.15 \\
Edge blur, $24$\,px                & $-0.18$ & $-0.50$ & 2.68 & 6.06 \\
\bottomrule
\end{tabular}}}
\label{tab:depth_failures}\vspace{-0.5em}
\end{table}

\vspace{1mm} \noindent \textbf{Depth Robustness.} Since reprojection depends on the estimated depth $D(p)$ in Equation~\ref{eq:unproject}, the sensitivity of CamTrol++ to the depth predictor and to depth errors is measured. Table~\ref{tab:depth_ablation} first compares ZoeDepth~\citep{bhat2023zoedepth}, DepthPro~\citep{bochkovskii2024depth}, DA3~\citep{lin2025depth}, and Marigold~\citep{ke2025marigold}. Metric depth from ZoeDepth gives the best perceptual and angular consistency, and the gap to WAVE narrows with the relative depth of Marigold ($0.2084$ against $0.2408$ LPIPS-\textit{next}), so depth quality matters for the final quality. Random corruption is then added to the ZoeDepth output before warping, in the form of relative Gaussian noise and missing pixels (Figure~\ref{fig:depth_vs_wave}, Table~\ref{tab:supp_depth_grid}). This sweep uses the $20$ MegaScenes images of the matched studies along a two-direction path that turns $30^\circ$ to each side of the input view in $15^\circ$ passes ($56$ frames), on which uncorrupted CamTrol++ reaches an LPIPS-\textit{next} of $0.099$. At severity $0.5$, the strongest noise tested, it rises to $0.115$, an increase of about $16\%$, and stays well below the $0.2408$ of WAVE on the larger MegaScenes set of Table~\ref{table:table_1}.

Table~\ref{tab:depth_failures} tests structured depth failures, which are harder than random noise because they do not average out during warping. Depth that collapses towards a plane, depth that is blurred across object edges and depth that is too far all keep the generated views consistent with the camera. A fully flat depth gives a planar warp, which agrees with the epipolar constraint by construction, so EAE alone does not describe this case. Its LPIPS-\textit{next} ($0.114$) at $15^\circ$ stays close to the default ($0.124$), and its ZMR ($0.085$) rises above the default ($0.058$) but stays far below the single pass ($0.234$), so the views keep moving. A depth scale that is too small is the failure that matters. For a sideways camera translation $t$, Equation~\ref{eq:project} shifts a pixel by roughly $f\,\|t\|/D(p)$ in addition to the shift caused by rotation, where $f$ is the focal length. A depth map scaled by a factor $s<1$ therefore enlarges every translation-induced shift by $1/s$, and with it the parallax of every warp. Small steps limit its effect to $+1.32\times10^{-3}$ EAE, against $+5.83\times10^{-3}$ for a single pass, and under every failure $15^\circ$ steps keep a much lower EAE than the single pass.

\section{Discussion}
\label{sec:discussion}

CamTrol++ demonstrates that camera-controlled novel view synthesis can be stabilized at inference time without retraining or modifying the diffusion backbone. The main source of this stability is a bound on the camera motion that each diffusion pass has to handle. Each pass then starts from a warp whose distortion and disoccluded area stay small, so the backbone can repair it without losing geometry or freezing the camera. The matched step-size study supports this design. Geometric error grows steadily with the arc per pass, and a single SVD pass is both less consistent and more prone to stalling. The same bound improves CamTrol when it is run with the same passes, improves 3D reconstruction for both SVD and LTX-Video, and keeps its advantage under every structured depth failure tested.

The bound also explains when splitting is needed. A path that rotates by less than the bound is generated in one pass, which keeps fidelity to real frames on short RealEstate10K paths, while a $60^\circ$ orbit is split into four passes, which more than halves the geometric error and lets COLMAP register all $20$ MegaScenes sequences ($12$ of $14$ on MipNeRF360). The best bound depends on the backbone. SVD, with its short $14$-frame window, benefits from small arcs, while LTX-Video covers larger arcs per pass and reaches its lowest epipolar error at $20^\circ$ to $30^\circ$. A bound on the arc per pass, rather than a fixed number of passes, is therefore the transferable part of the method.

Unlike image-centric approaches such as WAVE~\citep{park2025wave}, which use an image diffusion model without a temporal prior and may produce static or inconsistent motion under long trajectories, CamTrol++ maintains explicit geometric structure throughout the generation process. Epipolar attention gives a small geometric refinement within each pass, and anchoring at pass boundaries removes most of the color drift that would otherwise build up across passes. The registration-free warping makes the multi-pass design practical, since four CamTrol++ passes take about as long as a single CamTrol pass with registration ($222$\,s against $218$\,s per scene).

The long-horizon and depth robustness experiments show that the framework remains effective beyond the standard setting. CamTrol++ maintains temporal and geometric consistency over 56-frame sequences and remains effective under controlled depth corruption. At longer horizons, loss of texture and fine details becomes the dominant source of degradation. Extending generation further will therefore benefit from improvements in the underlying diffusion model and depth estimation in addition to inference-time stabilization.

Recent long-video stabilization methods address temporal drift in other ways. FreeLong~\citep{lu2024freelong} blends global and local spectral features of the video model without training, and Rolling Forcing~\citep{liu2025rolling} trains the model to denoise frames in a rolling window. Both target the length of a video rather than the size of the camera motion. In contrast, CamTrol++ improves camera-controlled multi-view generation entirely at inference time while preserving the original diffusion backbone. This makes the framework modular and applicable without additional model training.

From an application perspective, CamTrol++ enables longer camera-controlled visualizations from a single image using existing pre-trained video diffusion models. This is useful for applications such as interactive scene exploration, virtual walkthroughs, and 3D content creation, where consistent viewpoint changes are required without per-scene training. Its registration-free implementation and inference-time operation make it practical to integrate into existing image-to-video pipelines.

\noindent\textbf{Limitations.} Very long trajectories can still exhibit texture drift, loss of fine details, or depth inconsistencies. Performance also depends on the quality of monocular depth estimation and may degrade under severe occlusion or reflective surfaces. A depth scale that is too small enlarges the parallax of every warp and remains the most harmful depth error, although small steps limit its effect. Depth that is flattened or blurred keeps the views consistent with the camera, and its effect on the reconstructed shape is best assessed against real views. The bound on the arc per pass is set once per backbone, and selecting it automatically from the scene and the path is a natural extension. Future work could improve depth estimation, incorporate depth refinement, or use explicit 3D representations to further extend the generation horizon.
\section{Conclusion}
\label{sec:conclusion}

The work proposes CamTrol++, a training-free framework for stabilizing camera-controlled novel view synthesis at inference time. The key idea is to bound the camera rotation that each diffusion pass has to handle and to generate the path in as many passes as the bound requires. This limits per-pass geometric distortion and error accumulation without retraining or modifying the diffusion backbone. Geometry-constrained attention and lightweight perceptual anchoring provide additional consistency, while registration-free warping substantially reduces the cost of generation. On a matched $60^\circ$ orbit, CamTrol++ achieves the lowest epipolar error and the best held-out 3D reconstruction among CamTrol, NVS-Solver and WAVE, and it matches CamTrol on real RealEstate10K camera motion at half the run time (NVS-Solver is more faithful but $8\times$ slower). A matched step-size study shows that geometric error grows steadily with the arc per pass, and the same bound improves reconstruction with a second backbone, LTX-Video. These results demonstrate that careful control of camera motion at inference time can substantially improve the stability and efficiency of camera-controlled novel view synthesis while keeping the underlying diffusion model frozen.

\section*{Acknowledgment}
This work was supported by the Prime Minister Research Fellowship (PMRF 2122-2557) awarded to Prajwal Singh, the Visvesvaraya Fellowship awarded to Arjun Badola, and the Dr.~Vilas Mujumdar Chair held by Shanmuganathan Raman.

\bibliography{main}

@String(CVPR= {IEEE Conf. Comput. Vis. Pattern Recog.})

@String(ICCV= {Int. Conf. Comput. Vis.})

@String(ECCV= {Eur. Conf. Comput. Vis.})

@String(TOG= {ACM Trans. Graph.})

@String(ICLR = {Int. Conf. Learn. Represent.})

@String(CVPR  = {CVPR})

@String(ICCV  = {ICCV})

@String(ECCV  = {ECCV})

@String(TOG   = {ACM TOG})

@String(ICLR  = {ICLR})

@article{blattmann2023stable,
  title={{Stable Video Diffusion}: Scaling latent video diffusion models to large datasets},
  author={Blattmann, Andreas and Dockhorn, Tim and Kulal, Sumith and Mendelevitch, Daniel and Kilian, Maciej and Lorenz, Dominik and Levi, Yam and English, Zion and Voleti, Vikram and Letts, Adam and others},
  journal={arXiv preprint arXiv:2311.15127},
  year={2023}
}

@inproceedings{hou2024training,
  title={Training-free camera control for video generation},
  author={Hou, Chen and Chen, Zhibo},
  booktitle={International Conference on Learning Representations (ICLR)},
  year={2025}
}

@article{zhou2018stereo,
  journal={ACM Transactions on Graphics (TOG)},
  title={Stereo magnification: Learning view synthesis using multiplane images},
  author={Zhou, Tinghui and Tucker, Richard and Flynn, John and Fyffe, Graham and Snavely, Noah},
  volume={37},
  number={4},
  pages={65:1--65:12},
  year={2018},
  organization={ACM}
}

@inproceedings{tung2024megascenes,
  title={{MegaScenes}: Scene-level view synthesis at scale},
  author={Tung, Joseph and Chou, Gene and Cai, Ruojin and Yang, Guandao and Zhang, Kai and Wetzstein, Gordon and Hariharan, Bharath and Snavely, Noah},
  booktitle={European Conference on Computer Vision (ECCV)},
  pages={197--214},
  year={2024},
}

@inproceedings{park2025wave,
  title={{WAVE}: Warp-based view guidance for consistent novel view synthesis using a single image},
  author={Park, Jiwoo and Choi, Tae Eun and Jun, Youngjun and Hwang, Seong Jae},
  booktitle={Proceedings of the IEEE/CVF International Conference on Computer Vision (ICCV)},
  pages={11906--11915},
  year={2025}
}

@inproceedings{rombach2021high,
  title={High-Resolution Image Synthesis with Latent Diffusion Models},
  author={Rombach, Robin and Blattmann, Andreas and Lorenz, Dominik and Esser, Patrick and Ommer, Bj{\"o}rn},
  booktitle={Proceedings of the IEEE/CVF Conference on Computer Vision and Pattern Recognition (CVPR)},
  pages={10684--10695},
  year={2022}
}

@article{bhat2023zoedepth,
  title={{ZoeDepth}: Zero-shot transfer by combining relative and metric depth},
  author={Bhat, Shariq Farooq and Birkl, Reiner and Wofk, Diana and Wonka, Peter and M{\"u}ller, Matthias},
  journal={arXiv preprint arXiv:2302.12288},
  year={2023}
}

@inproceedings{zwicker2002surface,
  title={Surface splatting},
  author={Zwicker, Matthias and Pfister, Hanspeter and Van Baar, Jeroen and Gross, Markus},
  booktitle={Proceedings of the 28th Annual Conference on Computer Graphics and Interactive Techniques (SIGGRAPH)},
  pages={371--378},
  year={2001}
}

@article{danielsson1980euclidean,
  title={Euclidean distance mapping},
  author={Danielsson, Per-Erik},
  journal={Computer Graphics and Image Processing},
  volume={14},
  number={3},
  pages={227--248},
  year={1980},
}

@inproceedings{lu2024freelong,
  booktitle={Advances in Neural Information Processing Systems (NeurIPS)},
  title={{FreeLong}: Training-free long video generation with {SpectralBlend} temporal attention},
  author={Lu, Yu and Liang, Yuanzhi and Zhu, Linchao and Yang, Yi},
  pages={131434--131455},
  year={2024}
}

@inproceedings{sadat2024eliminating,
  title={Eliminating oversaturation and artifacts of high guidance scales in diffusion models},
  author={Sadat, Seyedmorteza and Hilliges, Otmar and Weber, Romann M},
  booktitle={International Conference on Learning Representations (ICLR)},
  year={2025}
}

@article{reinhard2002color,
  title={Color transfer between images},
  author={Reinhard, Erik and Ashikhmin, Michael and Gooch, Bruce and Shirley, Peter},
  journal={IEEE Computer Graphics and Applications},
  volume={21},
  number={5},
  pages={34--41},
  year={2001},
}

@article{chang1978optimal,
  title={Optimal histogram matching by monotone gray level transformation},
  author={Chang, Shi-Kuo and Wong, Yin-Wah},
  journal={Communications of the ACM},
  volume={21},
  number={10},
  pages={835--840},
  year={1978},
}

@inproceedings{ye2025synthesizing,
  title={Synthesizing consistent novel views via {3D} epipolar attention without re-training},
  author={Ye, Botao and Liu, Sifei and Li, Xueting and Pollefeys, Marc and Yang, Ming-Hsuan},
  booktitle={International Conference on 3D Vision (3DV)},
  year={2025}
}

@article{mildenhall2021nerf,
  title={{NeRF}: Representing scenes as neural radiance fields for view synthesis},
  author={Mildenhall, Ben and Srinivasan, Pratul P and Tancik, Matthew and Barron, Jonathan T and Ramamoorthi, Ravi and Ng, Ren},
  journal={Communications of the ACM},
  volume={65},
  number={1},
  pages={99--106},
  year={2021},
}

@inproceedings{barron2021mip,
  title={{Mip-NeRF}: A multiscale representation for anti-aliasing neural radiance fields},
  author={Barron, Jonathan T and Mildenhall, Ben and Tancik, Matthew and Hedman, Peter and Martin-Brualla, Ricardo and Srinivasan, Pratul P},
  booktitle={Proceedings of the IEEE/CVF International Conference on Computer Vision (ICCV)},
  pages={5855--5864},
  year={2021}
}

@article{muller2022instant,
  title={Instant neural graphics primitives with a multiresolution hash encoding},
  author={M{\"u}ller, Thomas and Evans, Alex and Schied, Christoph and Keller, Alexander},
  journal={ACM Transactions on Graphics (TOG)},
  volume={41},
  number={4},
  pages={102:1--102:15},
  year={2022},
}

@article{kerbl20233d,
  title={{3D} {G}aussian splatting for real-time radiance field rendering},
  author={Kerbl, Bernhard and Kopanas, Georgios and Leimk{\"u}hler, Thomas and Drettakis, George},
  journal={ACM Transactions on Graphics (TOG)},
  volume={42},
  number={4},
  pages={139:1--139:14},
  year={2023}
}

@InProceedings{Lu_2024_CVPR,
    author    = {Lu, Zhicheng and Guo, Xiang and Hui, Le and Chen, Tianrui and Yang, Min and Tang, Xiao and Zhu, Feng and Dai, Yuchao},
    title     = {{3D} geometry-aware deformable {G}aussian splatting for dynamic view synthesis},
    booktitle = {Proceedings of the IEEE/CVF Conference on Computer Vision and Pattern Recognition (CVPR)},
    year      = {2024},
    pages     = {8900--8910}
}

@inproceedings{zhang2025egogaussian,
  title={{EgoGaussian}: Dynamic scene understanding from egocentric video with {3D} {G}aussian splatting},
  author={Zhang, Daiwei and Li, Gengyan and Li, Jiajie and Bressieux, Micka{\"e}l and Hilliges, Otmar and Pollefeys, Marc and Van Gool, Luc and Wang, Xi},
  booktitle={International Conference on 3D Vision (3DV)},
  pages={1091--1102},
  year={2025},
}

@inproceedings{liu2023zero1to3zeroshotimage3d,
  title={{Zero-1-to-3}: Zero-shot one image to {3D} object},
  author={Liu, Ruoshi and Wu, Rundi and Van Hoorick, Basile and Tokmakov, Pavel and Zakharov, Sergey and Vondrick, Carl},
  booktitle={Proceedings of the IEEE/CVF International Conference on Computer Vision (ICCV)},
  pages={9298--9309},
  year={2023}
}

@article{shi2023zero123plus,
  title={{Zero123++}: A single image to consistent multi-view diffusion base model},
  author={Shi, Ruoxi and Chen, Hansheng and Zhang, Zhuoyang and Liu, Minghua and Xu, Chao and Wei, Xinyue and Chen, Linghao and Zeng, Chong and Su, Hao},
  journal={arXiv preprint arXiv:2310.15110},
  year={2023}
}

@article{yu2024viewcraftertamingvideodiffusion,
  title={{ViewCrafter}: Taming video diffusion models for high-fidelity novel view synthesis},
  author={Yu, Wangbo and Xing, Jinbo and Yuan, Li and Hu, Wenbo and Li, Xiaoyu and Huang, Zhipeng and Gao, Xiangjun and Wong, Tien-Tsin and Shan, Ying and Tian, Yonghong},
  journal={arXiv preprint arXiv:2409.02048},
  year={2024}
}

@article{han2025zero4d,
  title={{Zero4D}: Training-free {4D} video generation from single video using off-the-shelf video diffusion},
  author={Park, Jangho and Kwon, Taesung and Ye, Jong Chul},
  journal={arXiv preprint arXiv:2503.22622},
  year={2025}
}

@inproceedings{yu2024wonderjourney,
  title={{WonderJourney}: Going from anywhere to everywhere},
  author={Yu, Hong-Xing and Duan, Haoyi and Hur, Junhwa and Sargent, Kyle and Rubinstein, Michael and Freeman, William T and Cole, Forrester and Sun, Deqing and Snavely, Noah and Wu, Jiajun and others},
  booktitle={Proceedings of the IEEE/CVF Conference on Computer Vision and Pattern Recognition (CVPR)},
  pages={6658--6667},
  year={2024},
}

@article{liu2025rolling,
  title={{Rolling Forcing}: Autoregressive long video diffusion in real time},
  author={Liu, Kunhao and Hu, Wenbo and Xu, Jiale and Shan, Ying and Lu, Shijian},
  journal={arXiv preprint arXiv:2509.25161},
  year={2025}
}

@book{zisserman2004multiple,
  title={Multiple View Geometry in Computer Vision},
  author={Hartley, Richard and Zisserman, Andrew},
  edition={2nd},
  publisher={Cambridge University Press},
  year={2003}
}

@inproceedings{farneback2003two,
  title={Two-frame motion estimation based on polynomial expansion},
  author={Farneb{\"a}ck, Gunnar},
  booktitle={Scandinavian Conference on Image Analysis (SCIA)},
  pages={363--370},
  year={2003},
}

@inproceedings{teed2020raft,
  title={{RAFT}: Recurrent all-pairs field transforms for optical flow},
  author={Teed, Zachary and Deng, Jia},
  booktitle={European Conference on Computer Vision (ECCV)},
  pages={402--419},
  year={2020},
}

@inproceedings{zhang2018unreasonable,
  title={The unreasonable effectiveness of deep features as a perceptual metric},
  author={Zhang, Richard and Isola, Phillip and Efros, Alexei A and Shechtman, Eli and Wang, Oliver},
  booktitle={Proceedings of the IEEE/CVF Conference on Computer Vision and Pattern Recognition (CVPR)},
  pages={586--595},
  year={2018}
}

@inproceedings{radford2021learning,
  title={Learning transferable visual models from natural language supervision},
  author={Radford, Alec and Kim, Jong Wook and Hallacy, Chris and Ramesh, Aditya and Goh, Gabriel and Agarwal, Sandhini and Sastry, Girish and Askell, Amanda and Mishkin, Pamela and Clark, Jack and others},
  booktitle={International Conference on Machine Learning (ICML)},
  pages={8748--8763},
  year={2021},
}

@inproceedings{schoenberger2016sfm,
    author={Sch\"{o}nberger, Johannes Lutz and Frahm, Jan-Michael},
    title={Structure-from-Motion Revisited},
    booktitle={Proceedings of the IEEE/CVF Conference on Computer Vision and Pattern Recognition (CVPR)},
    year={2016},
}

@inproceedings{schoenberger2016mvs,
    author={Sch\"{o}nberger, Johannes Lutz and Zheng, Enliang and Pollefeys, Marc and Frahm, Jan-Michael},
    title={Pixelwise View Selection for Unstructured Multi-View Stereo},
    booktitle={European Conference on Computer Vision (ECCV)},
    year={2016},
}

@inproceedings{brooks2023instructpix2pix,
  title={{InstructPix2Pix}: Learning to follow image editing instructions},
  author={Brooks, Tim and Holynski, Aleksander and Efros, Alexei A},
  booktitle={Proceedings of the IEEE/CVF Conference on Computer Vision and Pattern Recognition (CVPR)},
  pages={18392--18402},
  year={2023}
}

@inproceedings{poole2022dreamfusion,
  title={{DreamFusion}: Text-to-{3D} using {2D} diffusion},
  author={Poole, Ben and Jain, Ajay and Barron, Jonathan T and Mildenhall, Ben},
  booktitle={International Conference on Learning Representations (ICLR)},
  year={2023}
}

@article{cui2025uniview,
  title={{UniView}: Enhancing novel view synthesis from a single image by unifying reference features},
  author={Cui, Haowang and Chen, Rui and Wang, Jiaze and Guo, Tao and Qin, Zheng},
  journal={arXiv preprint arXiv:2509.04932},
  year={2025}
}

@article{kwon2024tweediemix,
  title={{TweedieMix}: Improving multi-concept fusion for diffusion-based image/video generation},
  author={Kwon, Gihyun and Ye, Jong Chul},
  journal={arXiv preprint arXiv:2410.05591},
  year={2024}
}

@inproceedings{hertz2022prompt,
  title={Prompt-to-Prompt Image Editing with Cross Attention Control},
  author={Hertz, Amir and Mokady, Ron and Tenenbaum, Jay and Aberman, Kfir and Pritch, Yael and Cohen-Or, Daniel},
  booktitle={International Conference on Learning Representations (ICLR)},
  year={2023}
}

@inproceedings{feng2022training,
  title={Training-Free Structured Diffusion Guidance for Compositional Text-to-Image Synthesis},
  author={Feng, Weixi and He, Xuehai and Fu, Tsu-Jui and Jampani, Varun and Akula, Arjun and Narayana, Pradyumna and Basu, Sugato and Wang, Xin Eric and Wang, William Yang},
  booktitle={International Conference on Learning Representations (ICLR)},
  year={2023}
}

@article{chefer2023attend,
  title={Attend-and-excite: Attention-based semantic guidance for text-to-image diffusion models},
  author={Chefer, Hila and Alaluf, Yuval and Vinker, Yael and Wolf, Lior and Cohen-Or, Daniel},
  journal={ACM Transactions on Graphics (TOG)},
  volume={42},
  number={4},
  pages={148:1--148:10},
  year={2023},
}

@inproceedings{kim2024fifo,
  booktitle={Advances in Neural Information Processing Systems (NeurIPS)},
  title={{FIFO-Diffusion}: Generating infinite videos from text without training},
  author={Kim, Jihwan and Kang, Junoh and Choi, Jinyoung and Han, Bohyung},
  pages={89834--89868},
  year={2024}
}

@article{yang2024compositional,
  title={Compositional video generation as flow equalization},
  author={Yang, Xingyi and Wang, Xinchao},
  journal={arXiv preprint arXiv:2407.06182},
  year={2024}
}

@article{gomel2024diffusion,
  title={Diffusion-based attention warping for consistent {3D} scene editing},
  author={Gomel, Eyal and Wolf, Lior},
  journal={arXiv preprint arXiv:2412.07984},
  year={2024}
}

@inproceedings{agarwal2023star,
  title={{A-STAR}: Test-time attention segregation and retention for text-to-image synthesis},
  author={Agarwal, Aishwarya and Karanam, Srikrishna and Joseph, KJ and Saxena, Apoorv and Goswami, Koustava and Srinivasan, Balaji Vasan},
  booktitle={Proceedings of the IEEE/CVF International Conference on Computer Vision (ICCV)},
  pages={2283--2293},
  year={2023}
}

@inproceedings{zhu2025training,
  title={Training-free Geometric Image Editing on Diffusion Models},
  author={Zhu, Hanshen and Zhu, Zhen and Zhang, Kaile and Gong, Yiming and Liu, Yuliang and Bai, Xiang},
  booktitle={Proceedings of the IEEE/CVF International Conference on Computer Vision (ICCV)},
  pages={19130--19140},
  year={2025}
}

@article{sun2024dimensionx,
  title={{DimensionX}: Create any {3D} and {4D} scenes from a single image with controllable video diffusion},
  author={Sun, Wenqiang and Chen, Shuo and Liu, Fangfu and Chen, Zilong and Duan, Yueqi and Zhang, Jun and Wang, Yikai},
  journal={arXiv preprint arXiv:2411.04928},
  year={2024}
}

@article{chung2023luciddreamer,
  title={{LucidDreamer}: Domain-free generation of {3D} {G}aussian splatting scenes},
  author={Chung, Jaeyoung and Lee, Suyoung and Nam, Hyeongjin and Lee, Jaerin and Lee, Kyoung Mu},
  journal={arXiv preprint arXiv:2311.13384},
  year={2023}
}

@inproceedings{zhang2025spatialcrafter,
  title={{SpatialCrafter}: Unleashing the imagination of video diffusion models for scene reconstruction from limited observations},
  author={Zhang, Songchun and Xu, Huiyao and Guo, Sitong and Xie, Zhongwei and Bao, Hujun and Xu, Weiwei and Zou, Changqing},
  booktitle={Proceedings of the IEEE/CVF International Conference on Computer Vision (ICCV)},
  pages={27794--27805},
  year={2025}
}

@inproceedings{hollein2024viewdiff,
  title={{ViewDiff}: {3D}-consistent image generation with text-to-image models},
  author={H{\"o}llein, Lukas and Bo{\v{z}}i{\v{c}}, Alja{\v{z}} and M{\"u}ller, Norman and Novotny, David and Tseng, Hung-Yu and Richardt, Christian and Zollh{\"o}fer, Michael and Nie{\ss}ner, Matthias},
  booktitle={Proceedings of the IEEE/CVF Conference on Computer Vision and Pattern Recognition (CVPR)},
  pages={5043--5052},
  year={2024}
}

@article{ye2023consistent,
  title={{Consistent-1-to-3}: Consistent image to {3D} view synthesis via geometry-aware diffusion models},
  author={Ye, Jianglong and Wang, Peng and Li, Kejie and Shi, Yichun and Wang, Heng},
  journal={arXiv preprint arXiv:2310.03020},
  year={2023}
}

@inproceedings{elata2025novel,
  title={Novel view synthesis with pixel-space diffusion models},
  author={Elata, Noam and Kawar, Bahjat and Ostrovsky-Berman, Yaron and Farber, Miriam and Sokolovsky, Ron},
  booktitle={Proceedings of the IEEE/CVF Conference on Computer Vision and Pattern Recognition (CVPR)},
  pages={26756--26766},
  year={2025}
}

@article{xu2024camco,
  title={{CamCo}: Camera-controllable {3D}-consistent image-to-video generation},
  author={Xu, Dejia and Nie, Weili and Liu, Chao and Liu, Sifei and Kautz, Jan and Wang, Zhangyang and Vahdat, Arash},
  journal={arXiv preprint arXiv:2406.02509},
  year={2024}
}

@article{zheng2024cami2v,
  title={{CamI2V}: Camera-controlled image-to-video diffusion model},
  author={Zheng, Guangcong and Li, Teng and Jiang, Rui and Lu, Yehao and Wu, Tao and Li, Xi},
  journal={arXiv preprint arXiv:2410.15957},
  year={2024}
}

@article{dai2025fantasyworld,
  title={{FantasyWorld}: Geometry-consistent world modeling via unified video and {3D} prediction},
  author={Dai, Yixiang and Jiang, Fan and Wang, Chiyu and Xu, Mu and Qi, Yonggang},
  journal={arXiv preprint arXiv:2509.21657},
  year={2025}
}

@inproceedings{song2025worldforge,
  title={Taming Video Models for {3D} and {4D} Generation via Zero-Shot Camera Control},
  author={Song, Chenxi and Yang, Yanming and Zhao, Tong and Li, Ruibo and Zhang, Chi},
  booktitle={Proceedings of the IEEE/CVF Conference on Computer Vision and Pattern Recognition (CVPR)},
  year={2026}
}

@article{zhou2025stable,
  title={{Stable Virtual Camera}: Generative view synthesis with diffusion models},
  author={Zhou, Jensen and Gao, Hang and Voleti, Vikram and Vasishta, Aaryaman and Yao, Chun-Han and Boss, Mark and Torr, Philip and Rupprecht, Christian and Jampani, Varun},
  journal={arXiv preprint arXiv:2503.14489},
  year={2025}
}

@inproceedings{fujiwara2024style,
  title={{Style-NeRF2NeRF}: {3D} style transfer from style-aligned multi-view images},
  author={Fujiwara, Haruo and Mukuta, Yusuke and Harada, Tatsuya},
  booktitle={SIGGRAPH Asia 2024 Conference Papers},
  pages={1--10},
  year={2024}
}

@article{mildenhall2019local,
  title={Local light field fusion: Practical view synthesis with prescriptive sampling guidelines},
  author={Mildenhall, Ben and Srinivasan, Pratul P and Ortiz-Cayon, Rodrigo and Kalantari, Nima Khademi and Ramamoorthi, Ravi and Ng, Ren and Kar, Abhishek},
  journal={ACM Transactions on Graphics (TOG)},
  volume={38},
  number={4},
  pages={29:1--29:14},
  year={2019},
}

@article{lin2025depth,
  title={{Depth Anything 3}: Recovering the visual space from any views},
  author={Lin, Haotong and Chen, Sili and Liew, Jun Hao and Chen, Donny Y. and Li, Zhenyu and Shi, Guang and Feng, Jiashi and Kang, Bingyi},
  journal={arXiv preprint arXiv:2511.10647},
  year={2025}
}

@article{ke2025marigold,
  title={{Marigold}: Affordable adaptation of diffusion-based image generators for image analysis},
  author={Ke, Bingxin and Qu, Kevin and Wang, Tianfu and Metzger, Nando and Huang, Shengyu and Li, Bo and Obukhov, Anton and Schindler, Konrad},
  journal={arXiv preprint arXiv:2505.09358},
  year={2025}
}

@inproceedings{bochkovskii2024depth,
  title={{Depth Pro}: Sharp monocular metric depth in less than a second},
  author={Bochkovskii, Aleksei and Delaunoy, Ama{\"e}l and Germain, Hugo and Santos, Marcel and Zhou, Yichao and Richter, Stephan R. and Koltun, Vladlen},
  booktitle={International Conference on Learning Representations (ICLR)},
  year={2025}
}

@inproceedings{fridman2023scenescape,
  booktitle={Advances in Neural Information Processing Systems (NeurIPS)},
  title={{SceneScape}: Text-driven consistent scene generation},
  author={Fridman, Rafail and Abecasis, Amit and Kasten, Yoni and Dekel, Tali},
  pages={39897--39914},
  year={2023}
}

@inproceedings{li2026neomap,
  booktitle={European Conference on Computer Vision (ECCV)},
  title={{NeoMap}: Training-free novel-view synthesis from single images and videos},
  author={Li, Jinxi and Zhang, Tianyi and Yang, Yafei and Zhang, Zihui and Huang, Peng and Lin, Koon Wing Macgyver and Yang, Bo},
  year={2026}
}

@inproceedings{barron2022mip,
  title={{Mip-NeRF 360}: Unbounded anti-aliased neural radiance fields},
  author={Barron, Jonathan T and Mildenhall, Ben and Verbin, Dor and Srinivasan, Pratul P and Hedman, Peter},
  booktitle={Proceedings of the IEEE/CVF Conference on Computer Vision and Pattern Recognition (CVPR)},
  pages={5460--5469},
  year={2022},
}

@inproceedings{you2025nvssolver,
  title={{NVS-Solver}: Video Diffusion Model as Zero-Shot Novel View Synthesizer},
  author={You, Meng and Zhu, Zhiyu and Liu, Hui and Hou, Junhui},
  booktitle={International Conference on Learning Representations (ICLR)},
  year={2025}
}

@inproceedings{seo2024genwarp,
  title={{GenWarp}: Single Image to Novel Views with Semantic-Preserving Generative Warping},
  author={Seo, Junyoung and Fukuda, Kazumi and Shibuya, Takashi and Narihira, Takuya and Murata, Naoki and Hu, Shoukang and Lai, Chieh-Hsin and Kim, Seungryong and Mitsufuji, Yuki},
  booktitle={Advances in Neural Information Processing Systems (NeurIPS)},
  year={2024}
}

@inproceedings{huang2024epidiff,
  title={{EpiDiff}: Enhancing Multi-View Synthesis via Localized Epipolar-Constrained Diffusion},
  author={Huang, Zehuan and Wen, Hao and Dong, Junting and Wang, Yaohui and Li, Yangguang and Chen, Xinyuan and Cao, Yan-Pei and Liang, Ding and Qiao, Yu and Dai, Bo and Sheng, Lu},
  booktitle={Proceedings of the IEEE/CVF Conference on Computer Vision and Pattern Recognition (CVPR)},
  year={2024}
}

@inproceedings{kwak2024vivid,
  title={{ViVid-1-to-3}: Novel View Synthesis with Video Diffusion Models},
  author={Kwak, Jeong-gi and Dong, Erqun and Jin, Yuhe and Ko, Hanseok and Mahajan, Shweta and Yi, Kwang Moo},
  booktitle={Proceedings of the IEEE/CVF Conference on Computer Vision and Pattern Recognition (CVPR)},
  year={2024}
}

@article{hacohen2024ltx,
  title={{LTX-Video}: Realtime Video Latent Diffusion},
  author={HaCohen, Yoav and Chiprut, Nisan and Brazowski, Benny and Shalem, Daniel and Moshe, Dudu and Richardson, Eitan and Levin, Eran and Shiran, Guy and Zabari, Nir and Gordon, Ori and others},
  journal={arXiv preprint arXiv:2501.00103},
  year={2024}
}

@article{wan2025wan,
  title={Wan: Open and Advanced Large-Scale Video Generative Models},
  author={{Wan Team}},
  journal={arXiv preprint arXiv:2503.20314},
  year={2025}
}

@inproceedings{wang2025vggt,
  title={{VGGT}: Visual Geometry Grounded Transformer},
  author={Wang, Jianyuan and Chen, Minghao and Karaev, Nikita and Vedaldi, Andrea and Rupprecht, Christian and Novotny, David},
  booktitle={Proceedings of the IEEE/CVF Conference on Computer Vision and Pattern Recognition (CVPR)},
  year={2025}
}

@article{ye2025gsplat,
  title={gsplat: An Open-Source Library for {G}aussian Splatting},
  author={Ye, Vickie and Li, Ruilong and Kerr, Justin and Turkulainen, Matias and Yi, Brent and Pan, Zhuoyang and Seiskari, Otto and Ye, Jianbo and Hu, Jeffrey and Tancik, Matthew and Kanazawa, Angjoo},
  journal={Journal of Machine Learning Research},
  volume={26},
  number={34},
  pages={1--17},
  year={2025}
}
\bibliographystyle{tmlr}

\clearpage
\appendix
\section{Implementation Details}
\label{sec:supp_implementation}

\vspace{1mm} \noindent \textbf{CamTrol++ Framework.} Algorithm~\ref{algo:main_algo} outlines the complete inference pipeline used in CamTrol++. The notation follows Section~\ref{sec:method}. Given an input image $I_0$ and a camera trajectory of $L=(N-1)K+1$ world-to-camera poses $\{[R_m \,|\, t_m]\}_{m=0}^{L-1}$, the method generates a sequence of camera-consistent frames without any training or fine-tuning. The trajectory is processed in $K$ chunks of $N$ frames, one diffusion pass per chunk, and consecutive chunks share one frame. Each chunk is synthesized with the steps below.

\begin{itemize}
    \item \textbf{Depth Estimation.} The depth $D$ of the input frame $I_{\text{input}}$ of the chunk is estimated with the pretrained ZoeDepth model.
    \item \textbf{Unprojection and Warping.} $I_{\text{input}}$ is lifted to a 3D point cloud with $D$ and projected into the other $N-1$ poses of the chunk (Equations~\ref{eq:unproject} and~\ref{eq:project}), and the projected points are forward splatted with a $3\times3$ kernel (Equation~\ref{eq:splat}).
    \item \textbf{Hole Filling.} Small gaps caused by disocclusions are filled by copying the nearest valid pixel, found with a distance transform. Pixels more than four pixels away from any projected point are instead marked as holes, and the hole mask is widened by five pixels. These large holes and regions outside the source view are then filled with Stable Diffusion inpainting, and the LAB histogram of each inpainted frame is matched to $I_0$.
    \item \textbf{Latent Encoding.} The warped frames are encoded into the latent space of Stable Video Diffusion (SVD).
    \item \textbf{Epipolar-Guided Attention.} Epipolar attention is applied at the 23\textsuperscript{rd} decoder layer of SVD with weight $\alpha=0.9$, using $n_s=W/2$ epipolar key samples per query, where $W$ is the latent width (Equations~\ref{eq:fundamental} and~\ref{eq:epipolar_blend}).
    \item \textbf{Decoding and LAB Anchoring.} The last decoded RGB frame of the chunk, $I_{\text{last}}$, is matched to the LAB histogram of $I_0$ before it starts the next chunk, which limits perceptual drift across chunks.
    \item \textbf{Autoregressive Step.} $I_{\text{last}}$ becomes the input frame $I_{\text{input}}$ of the next chunk, enabling long-horizon synthesis without retraining.
\end{itemize}

This design maintains spatial consistency through depth-based warping and epipolar priors while enforcing temporal coherence across chunks through LAB anchoring.

\begin{algorithm}[H]
\caption{CamTrol++ Algorithm for multi-view synthesis.}
\KwIn{Input image $I_0$, camera trajectory $\{[R_m \,|\, t_m]\}_{m=0}^{L-1}$, intrinsics $\mathbf{K}$}
\KwOut{Camera-consistent sequence of $L=(N-1)K+1$ frames}
\BlankLine
$I_{\text{input}} \leftarrow I_0$\;
\For{each chunk $i=1,\dots,K$}{
1. Estimate depth $D$ of $I_{\text{input}}$ using ZoeDepth; \\
2. Unproject $I_{\text{input}}$ with $D$ to a 3D point cloud (Equation~\ref{eq:unproject}); \\
3. Project to the other $N-1$ poses of the chunk (Equation~\ref{eq:project}) and splat with a $3\times3$ kernel (Equation~\ref{eq:splat}); \\
4. Fill small gaps using nearest-neighbor assignment with a distance transform; \\
5. Fill large missing regions and outpaint using Stable Diffusion; \\
5b. Match the LAB histogram of the inpainted frames to $I_0$; \\
6. Encode warped frames into SVD latent; \\
7. Apply epipolar-guided spatial attention ($\alpha=0.9$, Equations~\ref{eq:fundamental} and~\ref{eq:epipolar_blend}); \\
8. Decode to RGB and match the LAB histogram of the last frame $I_{\text{last}}$ to $I_0$; \\
9. $I_{\text{input}} \leftarrow I_{\text{last}}$; \\
}
\label{algo:main_algo}
\end{algorithm}


\begin{table}[!t]
\centering
\caption{\textbf{Framework Parameters and Arc-Size Validation.} (a) Parameters of the CamTrol++ framework. (b) Exact values of the arc-size validation, with epipolar attention turned off throughout to isolate the reprojection mechanism, $n=20$ scenes per arc size, MegaScenes. In this earlier sweep the total arc and the frame spacing changed with the arc per pass, so rows are not on matched poses. See Table~\ref{tab:step_size_matched} for the matched study. This earlier protocol also normalizes EAE differently, so its values are comparable within the table only.}\vspace{-0.5em}
\begin{minipage}[t]{0.39\linewidth}
\centering
\resizebox{\linewidth}{!}{
\def\arraystretch{1.0}
\begin{tabular}{lll}
\toprule
\textbf{Symbol} & \textbf{Description} & \textbf{Value} \\
\midrule
$N$ & Frames per chunk & 14 \\
$M$ & Total frames & 28 / 56 / 84 \\
$\alpha$ & Epipolar attention weight & 0.9 \\
$\theta_{\max}$ & Rotation bound per pass & $15^\circ$ \\
$n_s$ & Epipolar key samples & $W/2$ \\
 & Depth model & ZoeDepth \\
 & GPU & L40S (48 GB) \\
\bottomrule
\end{tabular}}

\vspace{0.4em}
(a)
\end{minipage}\hfill
\begin{minipage}[t]{0.59\linewidth}
\centering
\resizebox{\linewidth}{!}{
\def\arraystretch{1.1}
\begin{tabular}{lcccc}
\toprule
\textbf{Arc (deg)} & \textbf{LPIPS-\textit{next}} & \textbf{EAE} ($\times10^{-3}$) & \textbf{Sampson} ($\times10^{-4}$) & \textbf{FPA} \\
\midrule
10                    & 0.1158 & 0.563 & 2844.7  & 0.856 \\
\textbf{15 (default)} & 0.1000 & 0.744 & 4714.0  & 0.856 \\
18                    & 0.1040 & 0.874 & 6388.2  & 0.875 \\
20                    & 0.1114 & 1.056 & 10052.0 & 0.877 \\
22                    & 0.1243 & 1.100 & 10273.0 & 0.880 \\
25                    & 0.1404 & 1.151 & 10504.2 & 0.894 \\
30                    & 0.1680 & 1.425 & 16066.9 & 0.897 \\
40                    & 0.2414 & 2.090 & 34802.5 & 0.889 \\
\bottomrule
\end{tabular}}

\vspace{0.4em}
(b)
\end{minipage}
\label{tab:supp_table_1}
\label{tab:supp_arc_size}
\vspace{-0.5em}
\end{table}

\vspace{1mm} \noindent \textbf{Framework Parameters.} Table~\ref{tab:supp_table_1}(a) summarizes the key parameters used in all SVD experiments. Each chunk has $N=14$ frames and rotates the camera by at most $15^\circ$, and sequences of $M=28$ or $56$ frames are generated depending on the dataset setting. Here $M$ counts the frames of all chunks, so the frame shared by two consecutive chunks is counted in both. Epipolar attention employs $\alpha=0.9$ and $n_s=W/2$ key samples. Depth is predicted using ZoeDepth (N model), and all experiments are performed on a single NVIDIA L40S GPU (48\,GB). These parameters are fixed for all datasets to ensure consistent evaluation across different camera trajectories and scenes.

\section{Additional Analysis}
\label{sec:supp_characterization}

This section provides additional results supporting the experiments of Section~\ref{sec:experiments}. It includes extended CamTrol~\citep{hou2024training} comparisons, component analysis, camera arc-size results, the number of passes on real paths, 3D reconstruction on MipNeRF360, color-anchoring placement, $\alpha$-sensitivity, the full depth-corruption grid and the inpainted area per arc.


\vspace{1mm} \noindent \textbf{Extended CamTrol Comparison.} The original CamTrol~\citep{hou2024training} is designed for short 14-frame sequences. To compare the methods over a longer trajectory, the full camera path is divided into two independent 14-frame segments and CamTrol is run on each segment separately. The two outputs are then joined into a 28-frame sequence, giving CamTrol the same segmented trajectory structure used by CamTrol++. Table~\ref{table:supp_table_5_28frames} reports results on RealEstate10K~\citep{zhou2018stereo} and MegaScenes~\citep{tung2024megascenes}. CamTrol++ achieves better temporal consistency and camera accuracy while reducing the run time of a $28$-frame sequence, with all stages included, from about $15.4$ minutes per scene with CamTrol to $4.5$ minutes. These times were measured in the $28$-frame setting of this table and are not comparable with the $60^\circ$ orbit times of Table~\ref{tab:merged_geometry_3dgs}. These results show the benefit of combining small-step generation with the registration-free warping pipeline.


\vspace{1mm} \noindent \textbf{Extended Component Analysis.} \textit{(a) Epipolar Attention Layer Placement.} Table~\ref{tab:supp_table_2} evaluates epipolar attention at different SVD decoder layers. Insertion at layer $2$ over-constrains low-level features and suppresses motion (ZMR $0.999$), and layers $20$ and $[17, 20, 23]$ leave more of the image static than layers $5$ and $8$. Layers $8$, $17$ and $23$ behave similarly, and layer $23$ is used in all experiments as it gives the lowest ZMR among them. \textit{(b) Color Matching and Interpolation Analysis.} Table~\ref{tab:merged_color_interp} (left) compares RGB, HSV, and LAB color spaces for appearance correction. LAB achieves the lowest LPIPS and highest CLIPSIM for both First and Next comparisons and is therefore used for appearance anchoring. Table~\ref{tab:merged_color_interp} (right) compares the linear interpolation used in CamTrol with the splatting-based warping of CamTrol++. The two methods provide comparable perceptual quality. Table~\ref{tab:supp_runtime} breaks down the time of one pass by stage. Removing registration and replacing interpolation with splatting makes warping $6.1\times$ faster ($147.7$ against $24.3$ s for the $13$ warped frames of a pass, about $11.4$ and $1.9$ s per frame), and a full pass including SVD refinement $3.2\times$ faster.


\vspace{1mm} \noindent \textbf{Arc-Size Validation.} Table~\ref{tab:supp_arc_size}(b) reports an earlier step-size sweep from $10^\circ$ to $40^\circ$ with $n=20$ scenes per setting. In that sweep the total arc, the spacing between consecutive frames and the translation per degree changed together with the arc per pass. Since LPIPS-\textit{next} falls whenever consecutive frames are closer together, even for a perfect generator, differences between rows mix the effect of the step size with the effect of a different path. They are kept here only for reference. The matched study of Table~\ref{tab:step_size_matched} removes this confound by fixing the orbit and the scored poses.


\begin{table}[!t]
\centering
\caption{\textbf{Alpha Sensitivity, Exact Values.} Mean $\pm$ SEM, $n=20$ scenes, MegaScenes. Earlier method of Table~\ref{tab:supp_arc_size}, with values comparable within the table only.}\vspace{-0.5em}
\resizebox{0.8\linewidth}{!}{
\def\arraystretch{1.1}
\begin{tabular}{lcccc}
\toprule
$\boldsymbol{\alpha}$ & \textbf{LPIPS-\textit{next}} & \textbf{EAE} ($\times10^{-3}$) & \textbf{Sampson} ($\times10^{-4}$) & \textbf{FPA} \\
\midrule
0.0                & $0.0945 \pm 0.0052$ & $0.723 \pm 0.040$ & $4908.7 \pm 763.5$ & $0.854 \pm 0.015$ \\
0.3                & $0.0948 \pm 0.0053$ & $0.724 \pm 0.043$ & $4998.8 \pm 886.7$ & $0.856 \pm 0.014$ \\
0.5                & $0.0947 \pm 0.0053$ & $0.723 \pm 0.042$ & $4945.8 \pm 840.7$ & $0.856 \pm 0.014$ \\
0.7                & $0.0945 \pm 0.0052$ & $0.724 \pm 0.043$ & $5026.8 \pm 847.9$ & $0.858 \pm 0.014$ \\
\textbf{0.9 (default)} & $0.0946 \pm 0.0052$ & $0.721 \pm 0.041$ & $4913.4 \pm 772.3$ & $0.856 \pm 0.014$ \\
1.0                & $0.0941 \pm 0.0052$ & $0.716 \pm 0.040$ & $4863.1 \pm 757.5$ & $0.855 \pm 0.014$ \\
\bottomrule
\end{tabular}}
\label{tab:supp_alpha_sensitivity}\vspace{-0.5em}
\end{table}

\vspace{1mm} \noindent \textbf{Number of Passes on Short Real Paths.} Table~\ref{tab:supp_passes_re10k} splits each of the $44$ RealEstate10K paths into a fixed number of passes, regardless of its rotation. Since the median path rotates by only $10^\circ$, one pass already stays within the $15^\circ$ bound, and additional passes regenerate from already generated frames. Fidelity to the real frames is highest with one pass and decreases steadily with the number of passes, both with reprojection only and with the full system. Warping alone forward-warps the real input directly to each target pose and fills holes with the nearest pixel, without inpainting or diffusion. On these short moves it scores the highest PSNR, since it keeps the real pixels exactly and synthesizes no new content, but it leaves stretched and smeared regions wherever the warp has holes, so it serves only as a reference for what the visible pixels allow. A single pass is also slightly better than the $15^\circ$ bound on these paths ($17.79$ against $17.64$\,dB), because the two settings differ only on the $15$ clips that rotate by more than $15^\circ$, and splitting does not help on these short real paths. The bound matters for large arcs such as the $60^\circ$ orbit of Table~\ref{tab:merged_geometry_3dgs}.

\begin{table}[!t]
\centering
\caption{\textbf{Number of Passes on RealEstate10K.} Mean over $44$ clips and six target frames per clip, compared with the real frames at the same poses. The last row uses the $15^\circ$ bound of Section~\ref{sec:method}. Passes have $13$ frames here so that every pass count from $1$ to $6$ ends on the target poses, which is why the six-pass row differs from the six $14$-frame passes of Table~\ref{tab:re10k_fidelity}.}\vspace{-0.5em}
{%
\resizebox{0.8\linewidth}{!}{
\def\arraystretch{1.1}
\begin{tabular}{lcccc}
\toprule
\textbf{Setting} & \textbf{PSNR} $\uparrow$ & \textbf{SSIM} $\uparrow$ & \textbf{LPIPS} $\downarrow$ & \textbf{PSNR, warped pixels} $\uparrow$ \\
\midrule
Input frame repeated            & 13.60 & 0.352 & 0.357 & 13.85 \\
Warp only, no diffusion         & 19.29 & 0.678 & 0.144 & 20.78 \\
\midrule
Reprojection only, 1 pass       & 17.86 & 0.575 & 0.179 & 18.66 \\
Reprojection only, 2 passes     & 17.39 & 0.560 & 0.181 & 18.23 \\
Reprojection only, 3 passes     & 16.96 & 0.536 & 0.203 & 17.81 \\
Reprojection only, 6 passes     & 15.62 & 0.464 & 0.289 & 16.28 \\
\midrule
Full CamTrol++, 1 pass          & 17.79 & 0.571 & 0.181 & 18.59 \\
Full CamTrol++, 2 passes        & 17.40 & 0.557 & 0.183 & 18.25 \\
Full CamTrol++, 3 passes        & 17.03 & 0.536 & 0.203 & 17.90 \\
Full CamTrol++, 6 passes        & 15.97 & 0.472 & 0.280 & 16.67 \\
\rowcolor[HTML]{EFEFEF}
Full CamTrol++, at most $15^\circ$ per pass & 17.64 & 0.568 & 0.179 & 18.53 \\
\bottomrule
\end{tabular}}}
\label{tab:supp_passes_re10k}\vspace{-0.5em}
\end{table}

\vspace{1mm} \noindent \textbf{3D Reconstruction on MipNeRF360.} Table~\ref{tab:supp_mipnerf360} repeats the held-out 3D Gaussian Splatting protocol of Table~\ref{tab:merged_geometry_3dgs} on the $14$ MipNeRF360 images along the same $60^\circ$ orbit. Splitting the orbit into $15^\circ$ passes raises held-out SSIM from $0.317$ to $0.444$, lowers LPIPS from $0.423$ to $0.296$ and lets COLMAP register $12$ of $14$ sequences instead of $3$. The same holds with reprojection only, which reaches slightly higher PSNR and SSIM than the full system ($17.88$ against $17.42$\,dB), in line with the similar geometric error of the two settings in Table~\ref{tab:step_size_matched}.

\begin{figure}[t]
\centering
\includegraphics[width=\linewidth]{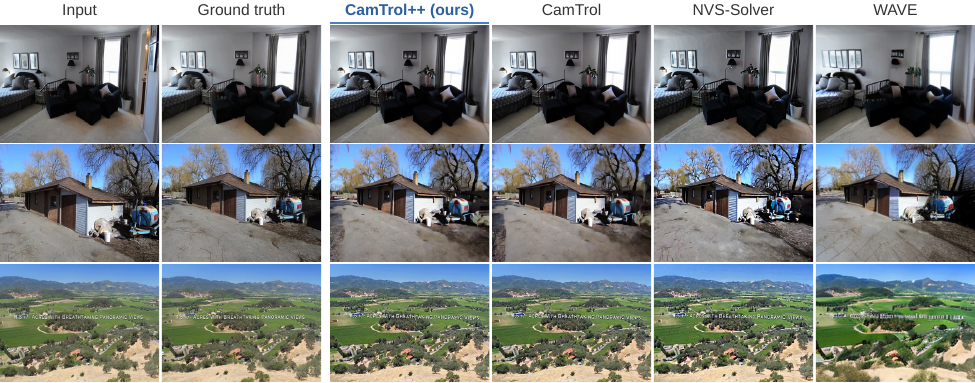}\vspace{-0.5em}
\caption{\textbf{Novel Views on RealEstate10K.} Each row shows the input frame, the real frame at one target pose, and the view generated by each method for that same pose. CamTrol++ matches the layout, colors and lighting of the ground truth closely in the first and third clips. CamTrol, which shares the same backbone, gives similar views on these short camera moves. NVS-Solver darkens the second clip and adds clutter around the trailer. WAVE brightens the room, bends the headboard and adds extra picture frames in the first clip, and blurs the overlaid text in the third clip.}
\label{fig:qual_re10k}
\end{figure}

\vspace{1mm} \noindent \textbf{Qualitative Comparison on RealEstate10K.} Figure~\ref{fig:qual_re10k} compares the views generated by each method with the real frames at the same target pose.

\begin{table}[!t]
\centering
\caption{\textbf{3D Reconstruction on MipNeRF360.} 3D Gaussian Splatting is fitted to each generated $60^\circ$ orbit with the commanded poses and evaluated on four held-out views, over $14$ MipNeRF360 images ($7$ scenes, $2$ views each). COLMAP counts the sequences for which all frames are registered.}\vspace{-0.5em}
{\small
\def\arraystretch{1.15}
\begin{tabular}{lcccc}
\toprule
\textbf{Setting} & \textbf{PSNR} $\uparrow$ & \textbf{SSIM} $\uparrow$ & \textbf{LPIPS} $\downarrow$ & \textbf{COLMAP} $\uparrow$ \\
\midrule
Reprojection only, single pass & 17.32 & 0.328 & 0.416 & 2/14 \\
Reprojection only, $15^\circ$ passes & \textbf{17.88} & \textbf{0.483} & \textbf{0.292} & 11/14 \\
\midrule
Full CamTrol++, single pass & 16.98 & 0.317 & 0.423 & 3/14 \\
\rowcolor[HTML]{EFEFEF}
Full CamTrol++, $15^\circ$ passes & 17.42 & 0.444 & 0.296 & \textbf{12/14} \\
\bottomrule
\end{tabular}}
\label{tab:supp_mipnerf360}\vspace{-0.5em}
\end{table}

\vspace{1mm} \noindent \textbf{Placement of Color Anchoring.} Table~\ref{tab:supp_color_placement} compares color anchoring placements against the real RealEstate10K frames, using the CIEDE2000 color difference on the pixels covered by the warp. Anchoring at pass boundaries gives the main gain and improves $39$ of $44$ clips. Adding the inpainted frames, as in CamTrol++, gives a similar error ($9.01$, $37$ of $44$ clips). Matching the inpainted frames alone slightly increases the error ($9.59$ against $9.53$), and a mean and variance transfer~\citep{reinhard2002color} at the boundaries performs on par with LAB histogram matching. The improvement does not depend on how much the real colors change along the path.

\begin{table}[!t]
\centering
\caption{\textbf{Placement of Color Anchoring.} Color difference to the real RealEstate10K frames on the pixels covered by the warp, over $44$ clips. Clips improved are counted against no anchoring.}\vspace{-0.5em}
{%
\resizebox{0.75\linewidth}{!}{
\def\arraystretch{1.1}
\begin{tabular}{lcc}
\toprule
\textbf{Color anchoring} & \textbf{CIEDE2000} $\downarrow$ & \textbf{Clips improved} \\
\midrule
None                                           & 9.53 & n/a \\
LAB on inpainted frames only                   & 9.59 & 16 of 44 \\
LAB at pass boundaries                         & 8.97 & 39 of 44 \\
\rowcolor[HTML]{EFEFEF}
LAB at pass boundaries and on inpainted frames (CamTrol++) & 9.01 & 37 of 44 \\
Mean and variance transfer at pass boundaries  & \textbf{8.94} & \textbf{40 of 44} \\
\bottomrule
\end{tabular}}}
\label{tab:supp_color_placement}\vspace{-0.5em}
\end{table}

\vspace{1mm} \noindent \textbf{Epipolar Attention $\alpha$-Sensitivity.} Table~\ref{tab:supp_alpha_sensitivity} shows the systematic study on $\alpha$ value. The blending weight is varied across $\alpha \in [0,1]$ using $n=20$ scenes. The results remain relatively stable across the tested values, showing that the overall generation quality is not strongly affected by the attention blend weight. Setting $\alpha=0$ disables the epipolar branch, and its values match those of $\alpha=0.9$ within the SEM, which agrees with the small gain of epipolar attention in Table~\ref{tab:merged_ablation_depth}. The default $\alpha=0.9$ and $\alpha=1.0$ give similar values. This sweep uses the earlier protocol of Table~\ref{tab:supp_arc_size}, whose EAE is normalized differently from the main text, so its values are comparable within the table only.





\begin{table}[!t]
\centering
\caption{\textbf{Depth Corruption Sensitivity, Exact Values.} $n=20$ scenes per cell, MegaScenes, full configuration, two-direction path of Section~\ref{sec:experiments} ($30^\circ$ to each side, $56$ frames), so values are not comparable with Table~\ref{tab:step_size_matched}. Uncorrupted LPIPS-\textit{next} on these scenes is $0.099$ and rises to at most $0.117$. The $56$-frame value of WAVE on the $98$ MegaScenes images of Table~\ref{table:table_1}, $0.2408$, is given only as a reference. Corresponds to Figure~\ref{fig:depth_heatmap}.}\vspace{-0.5em}
\resizebox{0.6\linewidth}{!}{
\def\arraystretch{1.05}
\begin{tabular}{cccccc}
\toprule
\textbf{Severity} & \textbf{Hole-frac} & \textbf{LPIPS-\textit{next}} & \textbf{LPIPS-\textit{first}} & \textbf{FPA} $\uparrow$ & \textbf{ZMR} $\downarrow$ \\
\midrule
0.00 & 0.00 & 0.0988 & 0.4629 & 0.811 & 0.157 \\
0.00 & 0.02 & 0.0994 & 0.4650 & 0.809 & 0.157 \\
0.00 & 0.05 & 0.0990 & 0.4618 & 0.812 & 0.161 \\
0.00 & 0.10 & 0.0989 & 0.4635 & 0.808 & 0.161 \\
\midrule
0.10 & 0.00 & 0.1026 & 0.4742 & 0.793 & 0.162 \\
0.10 & 0.02 & 0.1023 & 0.4729 & 0.793 & 0.160 \\
0.10 & 0.05 & 0.1029 & 0.4727 & 0.793 & 0.155 \\
0.10 & 0.10 & 0.1029 & 0.4701 & 0.791 & 0.156 \\
\midrule
0.20 & 0.00 & 0.1076 & 0.4771 & 0.766 & 0.153 \\
0.20 & 0.02 & 0.1079 & 0.4801 & 0.765 & 0.155 \\
0.20 & 0.05 & 0.1095 & 0.4812 & 0.759 & 0.145 \\
0.20 & 0.10 & 0.1079 & 0.4777 & 0.761 & 0.150 \\
\midrule
0.30 & 0.00 & 0.1093 & 0.4698 & 0.715 & 0.190 \\
0.30 & 0.02 & 0.1093 & 0.4714 & 0.713 & 0.196 \\
0.30 & 0.05 & 0.1113 & 0.4777 & 0.726 & 0.174 \\
0.30 & 0.10 & 0.1119 & 0.4765 & 0.715 & 0.183 \\
\midrule
0.50 & 0.00 & 0.1148 & 0.4612 & 0.693 & 0.228 \\
0.50 & 0.02 & 0.1146 & 0.4621 & 0.687 & 0.227 \\
0.50 & 0.05 & 0.1167 & 0.4633 & 0.689 & 0.219 \\
0.50 & 0.10 & 0.1164 & 0.4604 & 0.680 & 0.228 \\
\bottomrule
\end{tabular}}
\label{tab:supp_depth_grid}\vspace{-0.5em}
\end{table}

\begin{figure*}[!t]
\centering
\begin{minipage}[b]{0.95\linewidth}
  \centering
  \centerline{\includegraphics[width=\linewidth]{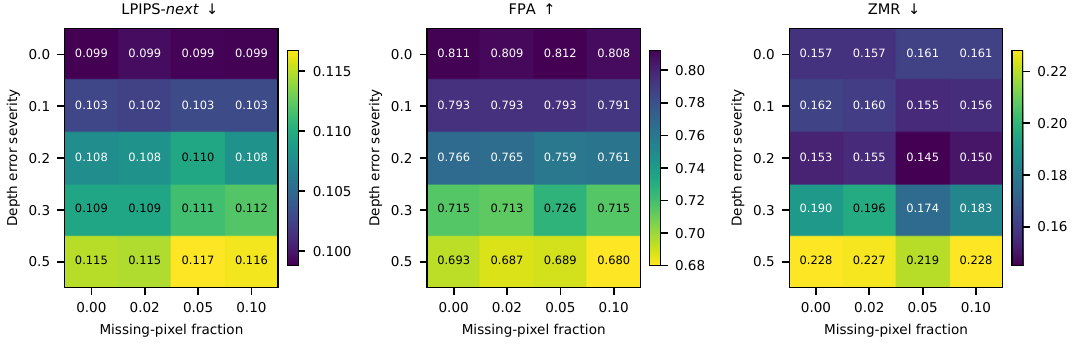}}
  \vspace{-2mm}
\end{minipage}
\captionof{figure}{\textbf{Depth Corruption Sensitivity Grid.} Full severity $\times$ hole-fraction sweep with $n=20$ scenes per cell on MegaScenes. Increasing depth-inaccuracy severity has a larger effect on LPIPS-\textit{next} than varying the missing-pixel fraction, which changes LPIPS-\textit{next} by less than $0.003$ at fixed severity.}
\label{fig:depth_heatmap}
\vspace{-0.5em}
\end{figure*}

\vspace{1mm} \noindent \textbf{Depth Corruption Sensitivity.} Table~\ref{tab:supp_depth_grid} and Figure~\ref{fig:depth_heatmap} provide the complete depth-corruption results underlying the depth robustness experiment of Section~\ref{sec:experiments} (Figure~\ref{fig:depth_vs_wave}). Both the severity of depth inaccuracy and the fraction of missing pixels are varied. Increasing depth error has a larger effect on performance than missing pixels across the tested range, consistent with the trend observed in the main text.

\begin{figure}[!t]
\centering
\includegraphics[width=0.55\linewidth]{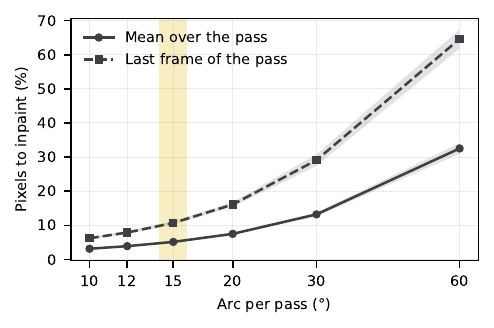}\vspace{-0.5em}
\caption{\textbf{Inpainted Area against Arc per Pass.} Fraction of each warped frame that is handed to Stable Diffusion inpainting, for the first pass of every setting of the matched step-size study. All settings warp the same input image with the same ZoeDepth depth and the commanded poses, so only the camera motion of the pass changes. Lines show the mean over $n=20$ MegaScenes images and shaded bands show $\pm1$ SEM. The solid line averages the $13$ warped frames of the pass and the dashed line shows its last frame. The yellow band marks the $15^\circ$ default.}
\label{fig:supp_hole_fraction}
\vspace{-0.5em}
\end{figure}

\vspace{1mm} \noindent \textbf{Inpainted Area and Arc per Pass.} Figure~\ref{fig:supp_hole_fraction} measures how much of each warped frame has to be inpainted as the arc per pass grows. The first pass of every setting of the matched study is warped from the same input image and depth with the commanded poses, using the warping of Section~\ref{sec:method}. A $15^\circ$ pass leaves $5.2\%$ of each warped frame to the inpainter on average and $10.7\%$ at its last frame, against $32.5\%$ and $64.6\%$ for a single $60^\circ$ pass. The area grows faster than the arc, from $3.2\%$ at $10^\circ$ to $13.2\%$ at $30^\circ$, and it is smaller at $15^\circ$ than at $60^\circ$ in all $20$ scenes. At the end of a single pass, the backbone therefore receives a frame that is mostly invented by the inpainter, which agrees with the loss of geometry and the frozen pixels of the single pass in Table~\ref{tab:step_size_matched}.

\section{Geometric and Motion Metrics}
\label{sec:supp_metrics}

\vspace{1mm} \noindent \textbf{Perceptual and Camera Metrics.} Perceptual and camera-based metrics follow the evaluation protocol of WAVE~\citep{park2025wave}. They include LPIPS~\citep{zhang2018unreasonable} and CLIPSIM \citep{radford2021learning} for temporal (Next) and reference (First) frame consistency, and Frobenius Norm, Rotation Angle Difference, and Angular Consistency for pose stability. Lower LPIPS, Frobenius Norm, Rotation Angle Difference, and Angular Consistency indicate better consistency, while higher CLIPSIM indicates better perceptual similarity.

\begin{table}[!t]
\centering
\caption{\textbf{Layer-wise Analysis of Epipolar Attention.} Epipolar attention is evaluated at different SVD decoder layers. Early-layer injection can strongly suppress motion, as shown by the high ZMR and low FPA at layer 2. Layers $8$, $17$ and $23$ behave similarly, and layer $23$, used in the final configuration, gives the lowest ZMR among them. This study uses the $56$-frame MegaScenes setting of Table~\ref{table:table_1} with a Sampson normalization that differs from the main text, so its Sampson values are comparable within this table only.}\vspace{-0.5em}
\resizebox{0.75\linewidth}{!}{
\def\arraystretch{1.3}
\begin{tabular}{lccccc}
\toprule
Injection Layer(s) & LPIPS-next $\downarrow$ & CLIPSIM-first $\uparrow$ & Sampson ($\times10^{-4}$) $\downarrow$ & FPA $\uparrow$ & ZMR $\downarrow$ \\
\midrule
\textit{\textbf{MegaScenes}}  &           &              &           &                      &       \\
\rowcolor[HTML]{FFDBBF}
2 & 0.0045 & 0.9885 & 0.14 & 0.376 & 0.999 \\
5 & 0.0762 & 0.8925 & 2.64 & 0.867 & 0.430 \\
8 & 0.0791 & 0.9137 & 2.74 & 0.894 & 0.409 \\
17 & 0.0753 & 0.9118 & 2.84 & 0.901 & 0.423 \\
20 & 0.0728 & 0.8859 & 2.77 & 0.835 & 0.481 \\
\rowcolor[HTML]{EFEFEF}
23 & 0.0797 & 0.9060 & 2.74 & 0.893 & 0.400 \\
{[5, 8]} & 0.0755 & 0.8890 & 2.68  & 0.866 & 0.445 \\
{[8, 23]} & 0.0785 & 0.9125 & 2.73  & 0.896 & 0.406 \\
{[17, 20, 23]} & 0.0675 & 0.8782 & 2.90 & 0.849 & 0.508\\
\bottomrule
\end{tabular}}
\label{tab:supp_table_2}\vspace{-1.0em}
\end{table}

\vspace{1mm} \noindent \textbf{Geometric Background.} A 3D point seen by two cameras projects to a pixel $x$ in the first image and a pixel $y$ in the second. With homogeneous coordinates $\tilde{x}=(x,1)^{\top}$ and $\tilde{y}=(y,1)^{\top}$, the two projections satisfy the epipolar constraint
\begin{align}
    \tilde{y}^{\top} F\, \tilde{x} = 0,
\end{align}
where the fundamental matrix $F$ encodes the relative rotation, translation and intrinsics of the two cameras up to scale~\citep{zisserman2004multiple}. Every point $x$ in the first view defines an epipolar line $\ell_{y}=F\tilde{x}$ in the second view, and the corresponding point $y$ lies exactly on that line if the geometry is perfect. In the metrics below, the two views are consecutive frames $I_t$ and $I_{t+1}$, and $F$ is built from their commanded poses as in Equation~\ref{eq:fundamental}, with frame $t+1$ in place of the reference frame $0$ and frame $t$ in place of frame $j$.

\begin{table*}[!t]
\centering
\captionof{table}{\textbf{Extended CamTrol Comparison.} CamTrol is evaluated with a split-trajectory strategy with two 14-frame segments to generate 28-frame sequences on RealEstate10K~\citep{zhou2018stereo} and MegaScenes~\citep{tung2024megascenes}. CamTrol++ provides stronger temporal and camera trajectory consistency, and the run times are given in Appendix~\ref{sec:supp_characterization}.}\vspace{-0.5em}
\resizebox{1.0\linewidth}{!}{
\def\arraystretch{1.2}
\begin{tabular}{llccccccccc}
\hline
\multicolumn{2}{c}{} &
  \multicolumn{2}{c}{Next} &
  \multicolumn{2}{c}{First} &
   &
   &
   &
   & \\ \cline{3-6}
\multicolumn{2}{c}{\multirow{-2}{*}{Method}} &
  LPIPS $\downarrow$ &
  CLIPSIM $\uparrow$ &
  LPIPS $\downarrow$ &
  CLIPSIM $\uparrow$ &
  \multirow{-2}{*}{\begin{tabular}[c]{@{}c@{}}Frob. Norm\\ (Rotation) $\downarrow$\end{tabular}} &
  \multirow{-2}{*}{\begin{tabular}[c]{@{}c@{}}Rot. Angle\\ Diff. $\downarrow$ \end{tabular}} &
  \multirow{-2}{*}{\begin{tabular}[c]{@{}c@{}}Angular \\ Cons. $\downarrow$\end{tabular}} &
  \multirow{-2}{*}{FPA $\uparrow$} &
  \multirow{-2}{*}{ZMR $\downarrow$} \\ \hline
\multicolumn{1}{l|}{} &
  \multicolumn{10}{l}{\textbf{\textit{RealEstate10K}}} \\
\multicolumn{1}{l|}{} &
  WAVE &
0.2602 &
0.9642 &
0.5457 &
0.9062 &
1.5917 &
1.1739 &
17.326 &
0.412 &
0.288 \\
\multicolumn{1}{l|}{} &
  CamTrol &
0.1896   &
0.9748   &
0.5194   &
0.9042   &
1.9505   &
1.5693   &
12.7253   &
0.804  &
\textbf{0.190}  \\
\multicolumn{1}{l|}{} &
  \cellcolor[HTML]{EFEFEF}CamTrol++ &
  \cellcolor[HTML]{EFEFEF}\textbf{0.1627} &
  \cellcolor[HTML]{EFEFEF}\textbf{0.9828} &
  \cellcolor[HTML]{EFEFEF}\textbf{0.5124} &
  \cellcolor[HTML]{EFEFEF}\textbf{0.9190} &
  \cellcolor[HTML]{EFEFEF}\textbf{1.2717} &
  \cellcolor[HTML]{EFEFEF}\textbf{1.0188} &
  \cellcolor[HTML]{EFEFEF}\textbf{8.6499} &
  \cellcolor[HTML]{EFEFEF}\textbf{0.844} &
  \cellcolor[HTML]{EFEFEF}0.200 \\ \cline{2-11}
\multicolumn{1}{l|}{} &
  \multicolumn{10}{l}{\textbf{\textit{MegaScenes}}} \\
\multicolumn{1}{l|}{} &
  WAVE & 
0.2872  &
0.9464  &
0.5367  &
0.8592  &
1.6963  &
\textbf{1.3086}  &
17.4329  &
0.502 &
0.297 \\
\multicolumn{1}{l|}{} &
  CamTrol &
0.1480  &
0.9793  &
0.4615  &
\textbf{0.9130 } &
1.7500  &
1.4910  &
15.2466  &
0.842  &
0.293 \\
\multicolumn{1}{l|}{\multirow{-8}{*}{\rotatebox[origin=c]{90}{28 Frames}}} &
  \cellcolor[HTML]{EFEFEF}CamTrol++ &
  \cellcolor[HTML]{EFEFEF}\textbf{0.1408} &
  \cellcolor[HTML]{EFEFEF}\textbf{0.9802} &
  \cellcolor[HTML]{EFEFEF}\textbf{0.4403} &
  \cellcolor[HTML]{EFEFEF}0.9101 &
  \cellcolor[HTML]{EFEFEF}\textbf{1.6517} &
  \cellcolor[HTML]{EFEFEF}1.3679 &
  \cellcolor[HTML]{EFEFEF}\textbf{8.6465} &
  \cellcolor[HTML]{EFEFEF}\textbf{0.847} &
  \cellcolor[HTML]{EFEFEF}\textbf{0.288} \\ \hline
\end{tabular}}
\label{table:supp_table_5_28frames} \vspace{-0.5em}
\end{table*}

\begin{table*}[!t]
\centering
\caption{\textbf{Color Matching and Interpolation Analysis.} (Left) Effect of different color spaces for appearance correction. LAB provides the best overall perceptual consistency among the tested choices. (Right) Comparison between the linear interpolation used in CamTrol~\citep{hou2024training} and the splatting-based warping of CamTrol++. Perceptual quality remains comparable. Time is the full warping stage per warped frame (depth, warping, hole filling and inpainting) at $960\times704$ on one L40S GPU, taken from the stage timing in Table~\ref{tab:supp_runtime}. Removing registration and replacing interpolation with splatting makes this stage $6.1\times$ faster.}\vspace{-0.5em}
\begin{minipage}{0.46\textwidth}
\centering
{\small
\resizebox{\linewidth}{!}{
\def\arraystretch{1.2}
\begin{tabular}{lcccc}
\hline
\multirow{2}{*}{Color Model} & \multicolumn{2}{c}{First} & \multicolumn{2}{c}{Next} \\ \cline{2-5}
 & LPIPS $\downarrow$ & CLIPSIM $\uparrow$ & LPIPS $\downarrow$ & CLIPSIM $\uparrow$ \\ 
\hline
\multicolumn{5}{l}{\textit{\textbf{MegaScenes}}} \\
RGB & 0.4382 & 0.9001 & 0.0800 & 0.9875 \\
HSV & 0.4456 & 0.8972 & 0.0817 & 0.9872 \\
\rowcolor[HTML]{EFEFEF}
LAB & \textbf{0.4340} & \textbf{0.9060} & \textbf{0.0796} & \textbf{0.9880} \\
\hline
\end{tabular}
}}
\end{minipage}
\hfill
\begin{minipage}{0.52\textwidth}
\centering
{\small
\resizebox{\linewidth}{!}{
\def\arraystretch{1.2}
\begin{tabular}{lccccc}
\hline
\multirow{2}{*}{Warping} & \multicolumn{2}{c}{First} & \multicolumn{2}{c}{Next} & \multirow{2}{*}{Time (s/frame) $\downarrow$} \\ \cline{2-5}
 & LPIPS $\downarrow$ & CLIPSIM $\uparrow$ & LPIPS $\downarrow$ & CLIPSIM $\uparrow$ & \\ 
\hline
\multicolumn{6}{l}{\textit{\textbf{MegaScenes}}} \\
CamTrol &  0.4340 & 0.9067 & 0.0800 & 0.9879 & 11.4 \\
\rowcolor[HTML]{EFEFEF}
CamTrol++ & 0.4340 & 0.9060 & 0.0796 & 0.9880 & \textbf{1.9} \\
\hline
\end{tabular}
}}
\end{minipage}
\label{tab:merged_color_interp}\vspace{-0.5em}
\end{table*}

\vspace{1mm} \noindent \textbf{Epipolar Alignment Error (EAE).} EAE measures how well correspondences between two views follow the epipolar geometry. For each correspondence $(x, y)$, it averages the distances of the two points to their epipolar lines,

\begin{equation}
\begin{split}
d_{\text{EAE}}(x, y; F) &= \frac{1}{2} \Bigg(
\frac{|\tilde{y}^{\top} F \tilde{x}|}{\sqrt{(F\tilde{x})_{1}^{2} + (F\tilde{x})_{2}^{2}}} \\
&+\frac{|\tilde{y}^{\top} F \tilde{x}|}{\sqrt{(F^{\top}\tilde{y})_{1}^{2} + (F^{\top}\tilde{y})_{2}^{2}}} \Bigg),
\end{split}
\end{equation}
where $(\cdot)_{i}$ is the $i$-th entry of a vector.
Lower values indicate that the corresponding points lie closer to their epipolar lines, implying stronger geometric alignment between views.  
All values are divided by the image diagonal, so EAE does not depend on the image size.

\vspace{1mm} \noindent \textbf{Sampson Distance.} The Sampson Distance provides a first-order approximation of the true reprojection error~\citep{zisserman2004multiple}. It refines the basic epipolar constraint by accounting for how sensitive the constraint is to small coordinate changes,

\begin{equation}
\begin{split}
d_{\text{Sampson}}(x, y; F) =
(\tilde{y}^{\top} F \tilde{x})^{2}
\bigg((F\tilde{x})_{1}^{2} + (F\tilde{x})_{2}^{2} + \\ (F^{\top}\tilde{y})_{1}^{2} + (F^{\top}\tilde{y})_{2}^{2}\bigg)^{-1}.
\end{split}
\end{equation}

\begin{figure*}[!t]
\centering
\begin{minipage}[b]{0.95\linewidth}
  \centering
  \centerline{\includegraphics[width=\linewidth]{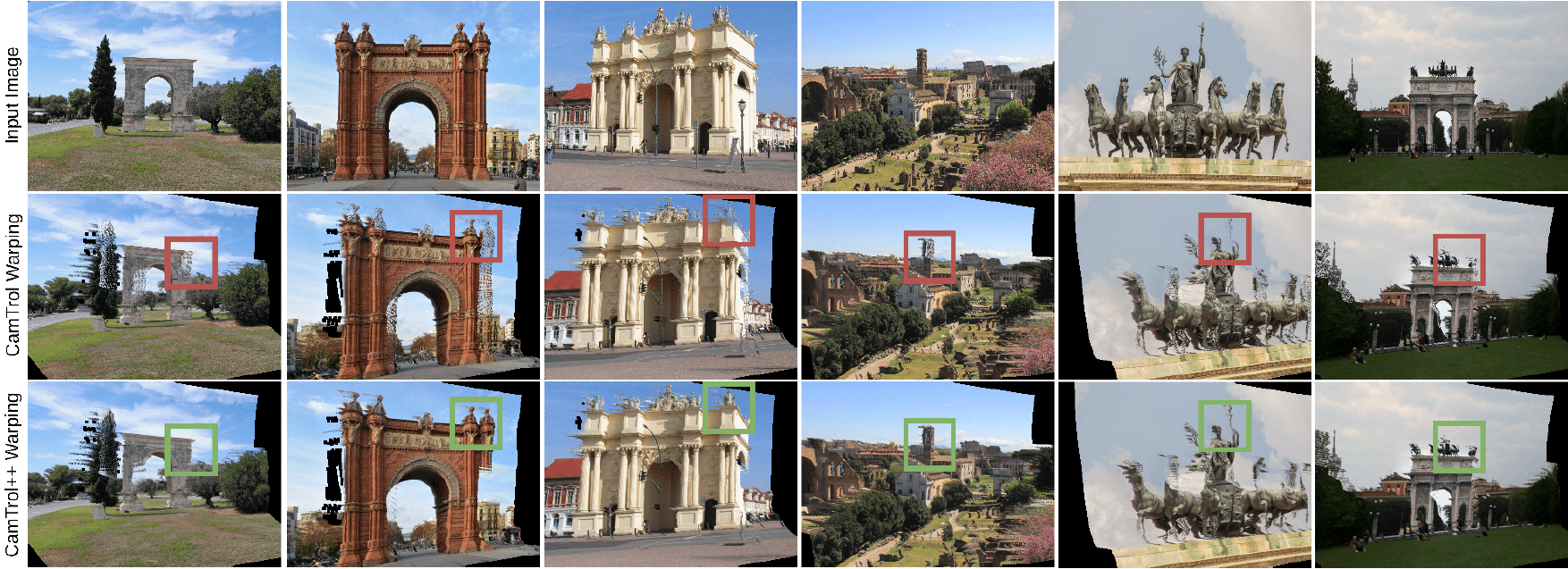}}
  \vspace{-2mm}
\end{minipage}
\captionof{figure}{\textbf{Warping Effect.} The figure shows the effect of using the linear interpolation-based warping method used by CamTrol~\citep{hou2024training} and the forward splatting-based warping of CamTrol++.}
\label{fig:figure_7_warpeffect} \vspace{-0.5em}
\end{figure*}
\begin{figure*}[!t]
\centering
\begin{minipage}[b]{0.95\linewidth}
  \centering
  \centerline{\includegraphics[width=\linewidth]{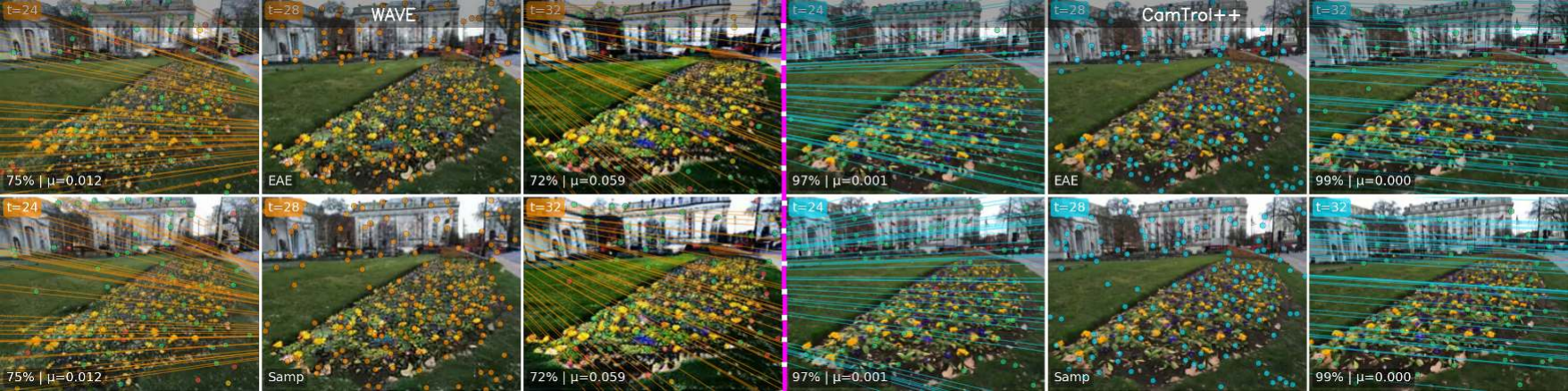}}
  \vspace{-2mm}
\end{minipage}
\captionof{figure}{\textbf{Geometric Consistency Visualization.} Feature tracks between frames are shown to assess geometric stability. \textit{Left, WAVE.} The feature tracks (orange) are geometrically inconsistent, as shown by the chaotic lines. \textit{Right, CamTrol++.} The feature tracks (cyan) are consistent and parallel and respect the perspective of the scene. Table~\ref{tab:merged_geometry_3dgs} gives the corresponding epipolar error over all scenes.}
\label{fig:figure_5_geometry} \vspace{-0.5em}
\end{figure*}

It measures the minimum perturbation required to satisfy the epipolar constraint. Lower values correspond to better pose consistency between the two views. In both metrics $F$ is not estimated from the images. It is computed from the commanded camera poses and the intrinsics rescaled to the $256\times192$ scoring size. Correspondences come from a regular grid of points every $12$ pixels in $I_t$, tracked to $I_{t+1}$ with RAFT flow. A point is kept only if an independent Farneb\"ack flow~\citep{farneback2003two} between the two frames maps it back to within $1\%$ of the longer image side (about $2.6$ pixels). For each frame pair the median over the kept points is taken, and the result is averaged over all consecutive frame pairs. Pairs without camera translation have no epipolar geometry and are skipped. Sampson Distance is in squared pixels, so its values can only be compared at the same resolution.

\vspace{1mm} \noindent \textbf{Flow Persistence Agreement (FPA).} FPA and the Zero-Motion Ratio (ZMR) are computed from dense optical flow estimated with RAFT~\citep{teed2020raft}. FPA does not use the camera poses. It measures how smoothly the motion changes over time, by comparing the flow of one frame pair with the flow of the next. For a sequence of $T$ frames, let $\mathbf{u}_{t}$ denote the flow from $I_{t}$ to $I_{t+1}$. Two consistency measures are computed,

\begin{align}
\text{FPA}_{\text{dir}} &= \frac{1}{T-2} \sum_{t=1}^{T-2}
\frac{1}{|\Omega|}\sum_{p \in \Omega}
\frac{\mathbf{u}_{t}(p) \cdot \mathbf{u}_{t+1}(p)}{\|\mathbf{u}_{t}(p)\|_{2}\,\|\mathbf{u}_{t+1}(p)\|_{2} + \epsilon}, \\
\text{FPA}_{\text{mag}} &= 1 - \frac{1}{T-2} \sum_{t=1}^{T-2}
\frac{1}{|\Omega|}\sum_{p \in \Omega}
\frac{\big|\|\mathbf{u}_{t}(p)\|_{2} - \|\mathbf{u}_{t+1}(p)\|_{2}\big|}{\|\mathbf{u}_{t}(p)\|_{2} + \|\mathbf{u}_{t+1}(p)\|_{2} + \epsilon}.
\end{align}

\begin{table}[!t]
\centering
\caption{\textbf{Stage Timing.} Mean time of one $14$-frame pass ($13$ warped frames, $15^\circ$ arc) over three MegaScenes images at $960\times704$ on one NVIDIA L40S GPU. Both pipelines use the same depth, inpainting and SVD models, and almost all of the difference comes from projection, registration and interpolation. Warping is $6.1\times$ faster ($11.4$ against $1.9$ s per warped frame) and a full pass is $3.2\times$ faster.}\vspace{-0.5em}
{\small
\def\arraystretch{1.15}
\begin{tabular}{lrr}
\toprule
\textbf{Stage (seconds per pass)} & \textbf{CamTrol} & \textbf{CamTrol++} \\
\midrule
Depth & 1.20 & 0.09 \\
Projection, masks and point cloud registration & 55.91 & 1.27 \\
Interpolation (CamTrol) or splatting and fill (CamTrol++) & 71.75 & 3.73 \\
Inpainting & 18.82 & 19.20 \\
\midrule
Warping stage, total & 147.68 & 24.30 \\
SVD refinement (encode, denoise, decode) & 30.94 & 30.94 \\
\midrule
Full pass & 178.62 & 55.24 \\
\bottomrule
\end{tabular}}
\label{tab:supp_runtime}\vspace{-0.5em}
\end{table}

\begin{figure*}[!t]
\centering
\begin{minipage}[b]{0.95\linewidth}
  \centering
  \centerline{\includegraphics[width=\linewidth]{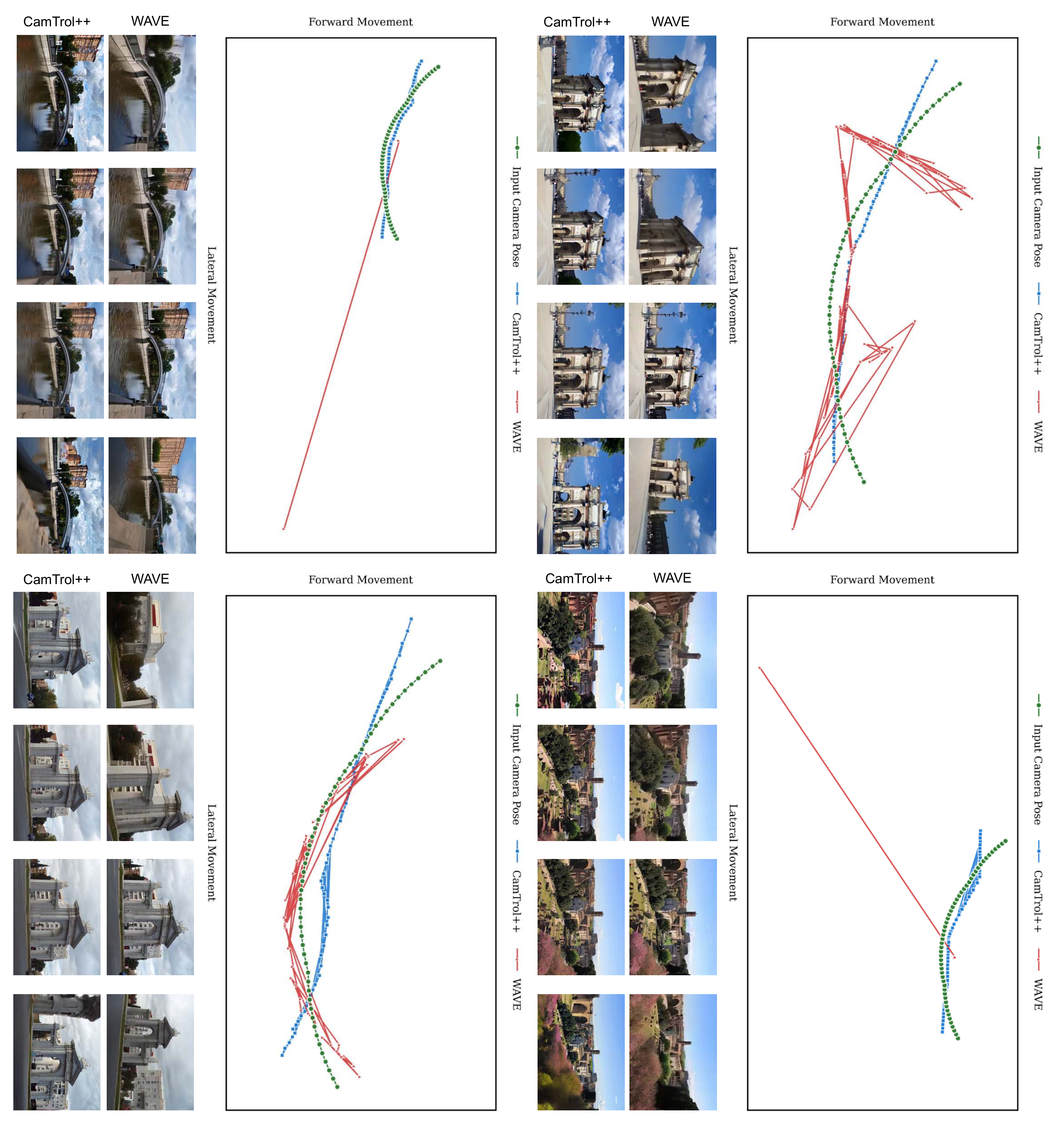}}
  \vspace{-1.2em}
\end{minipage}
\captionof{figure}{\textbf{Camera Trajectory.} Comparison between the ground-truth (GT) camera trajectory and the predicted camera poses estimated from generated videos using COLMAP~\citep{schoenberger2016sfm,schoenberger2016mvs}. Each curve shows the normalized translation parameters ($x$, $z$) for a scene from MegaScenes~\citep{tung2024megascenes}. For clarity, height variation is omitted as all poses are normalized to a common scale. CamTrol++ produces trajectories that align more closely with the ground truth, showing stable and consistent camera motion.}
\label{fig:supp_camera_param}
\vspace{-1.0em}
\end{figure*}
\begin{figure*}[!t]
\centering
\begin{minipage}[b]{1.0\linewidth}
  \centering
  \centerline{\includegraphics[width=\linewidth]{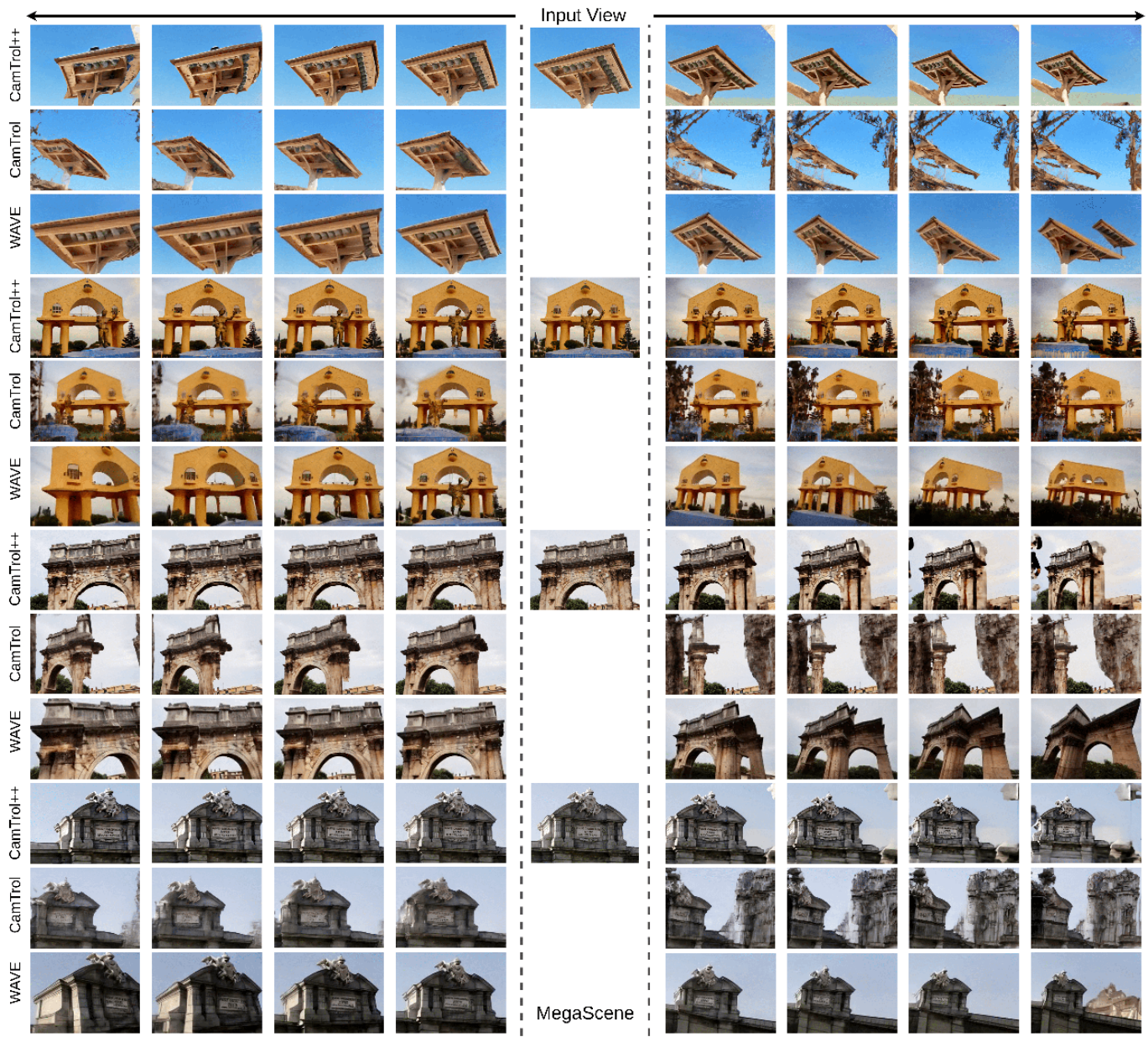}}
  \vspace{-0.5em}
\end{minipage}
\captionof{figure}{\textbf{Qualitative Result.} More qualitative results on the MegaScenes dataset~\citep{tung2024megascenes}. CamTrol++, CamTrol and WAVE follow the same path in both directions from the input view (center).}
\label{fig:supp_qual_megascene100}
\vspace{-2.0em}
\end{figure*}
\begin{figure*}[!t]
\centering
\begin{minipage}[b]{1.0\linewidth}
  \centering
  \centerline{\includegraphics[width=\linewidth]{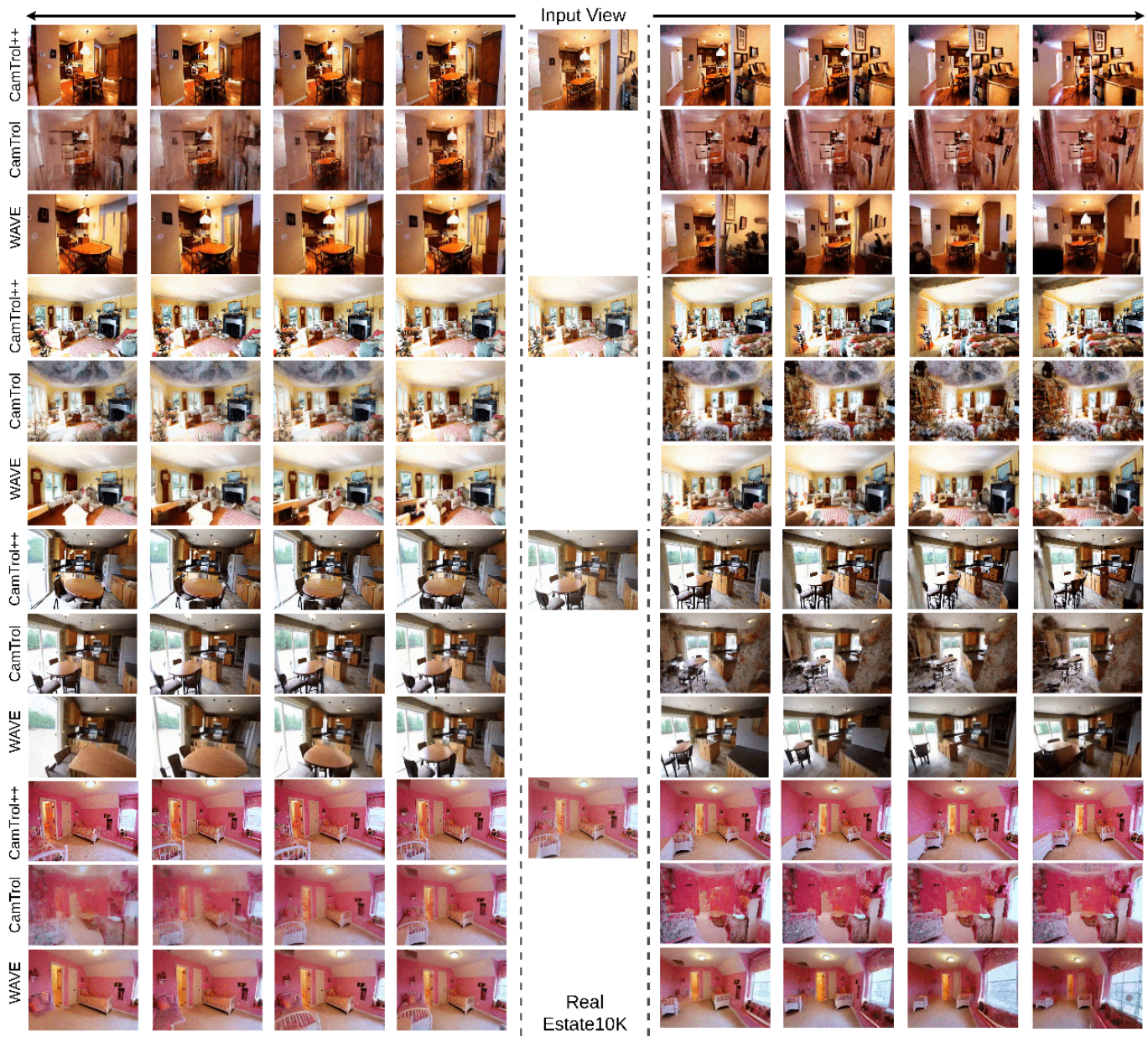}}
  \vspace{-0.5em}
\end{minipage}
\captionof{figure}{\textbf{Qualitative Result.} More qualitative results on RealEstate10K~\citep{zhou2018stereo}.}
\label{fig:supp_qual_realestate10k}
\vspace{-1.0em}
\end{figure*}

Here, $\Omega$ denotes the set of all pixel coordinates (the image domain), and $|\Omega|$ is the total number of pixels used to normalize the average across the frame. Directional consistency ($\text{FPA}_{\text{dir}}$) checks whether motion directions stay consistent, and magnitude consistency ($\text{FPA}_{\text{mag}}$) ensures the overall motion strength is smooth. The final score is the mean of both terms,

\begin{align}
\text{FPA} = \tfrac{1}{2}(\text{FPA}_{\text{dir}} + \text{FPA}_{\text{mag}}).
\end{align}

A higher FPA (close to 1) indicates stable and coherent motion, while a lower value suggests flicker or inconsistent movement across time.

\begin{figure*}[!t]
\centering
\begin{minipage}[b]{1.0\linewidth}
  \centering
  \centerline{\includegraphics[width=\linewidth]{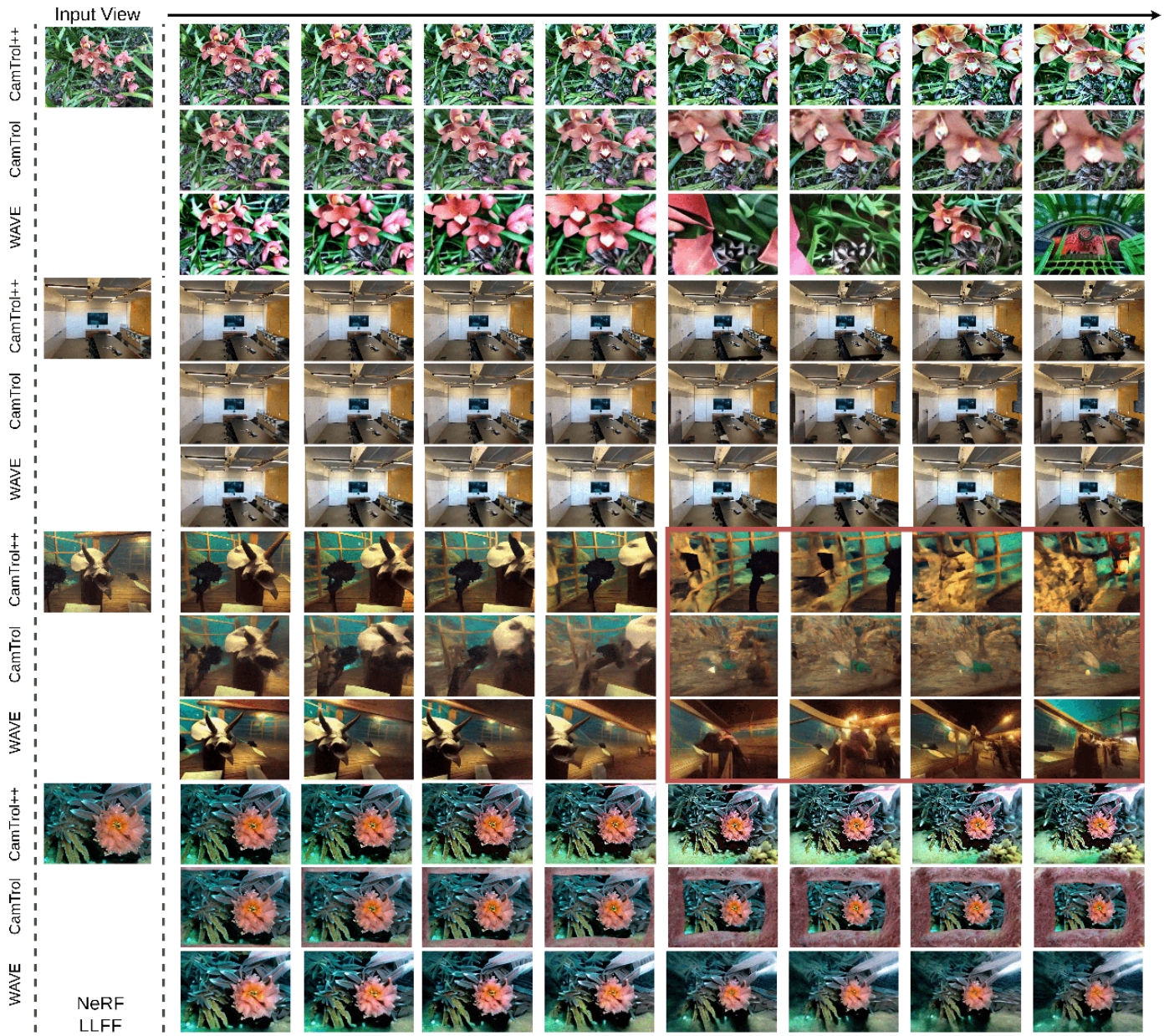}}
  \vspace{-0.5em}
\end{minipage}
\captionof{figure}{\textbf{Qualitative Result.} Additional results on the NeRF-LLFF dataset~\citep{mildenhall2019local} with complex camera motion and scene stylization. Across different scenes, CamTrol++ maintains sharper structure and more stable motion than CamTrol~\citep{hou2024training} and WAVE~\citep{park2025wave}. The third scene (red box) is a failure case, in which depth ambiguity may cause noticeable distortion across all training-free methods.
}
\label{fig:supp_qual_nerfllff}
\vspace{-1.0em}
\end{figure*}

\vspace{1mm} \noindent \textbf{Zero-Motion Ratio (ZMR).} ZMR measures how much of the image remains nearly static between frames. For the flow $\mathbf{u}_{t}$ from $I_{t}$ to $I_{t+1}$,

\begin{align}
\text{ZMR} = \frac{1}{|\Omega|} \sum_{p \in \Omega} \mathbf{1}\big(\|\mathbf{u}_{t}(p)\|_{2} < \tau\big),
\end{align}

where $\tau=0.75$ pixels at the $256\times192$ scoring size. A high ZMR means large regions have little to no motion (frozen frames), while a low ZMR indicates healthy camera movement. ZMR is averaged across all frame pairs, and together with FPA, provides a camera-agnostic measure of motion realism.

\vspace{1mm} \noindent \textbf{Chroma Drift and Saturation Deviation.} These two color metrics of Table~\ref{tab:merged_ablation_depth} are computed in CIELAB. Chroma drift is the distance between the mean $(a^*, b^*)$ of each frame and that of the first frame, averaged over all frames. Saturation deviation is $|\bar{C}_{\text{last}}/\bar{C}_{\text{first}} - 1|$, where $\bar{C}$ is the mean chroma $\sqrt{a^{*2}+b^{*2}}$ of a frame, so a value of $0$ means that the last frame keeps the saturation of the first.

\vspace{1mm} \noindent \textbf{Summary.} EAE and Sampson Distance evaluate geometric consistency against the commanded camera path, while FPA and ZMR assess motion directly from the video. Together, these metrics capture both spatial and temporal stability of generated sequences without requiring ground-truth 3D geometry.

\section{Camera Parameter Visualization}
\label{sec:supp_camera}

This section visualizes the recovered camera trajectories used in the camera accuracy evaluation of Table~\ref{table:table_1}. For each generated sequence, camera poses are estimated with COLMAP~\citep{schoenberger2016sfm,schoenberger2016mvs} and compared with the corresponding ground-truth orbital camera trajectory. Figure~\ref{fig:supp_camera_param} shows the normalized camera translations in the $x$-$z$ plane for representative scenes from MegaScenes~\citep{tung2024megascenes}. Since the poses are scale-normalized, vertical variation is minimal and therefore excluded for clarity following WAVE~\citep{park2025wave}. In Figure~\ref{fig:supp_camera_param}, CamTrol++ closely follows the ground-truth trajectory, indicating that the generated views maintain consistent motion across frames. In contrast, trajectories obtained from WAVE~\citep{park2025wave} exhibit greater deviation and irregular curvature, reflecting instability in the estimated camera poses. These examples agree with the lower Angular Consistency of CamTrol++ at $56$ frames in Table~\ref{table:table_1}. On MegaScenes, WAVE still has lower Frobenius norm and rotation angle errors.

\section{Additional Results and Details}
\label{sec:add_results}

The multi-view sequences generated by CamTrol++ are coherent and consistent. Additional results are shown in Figures~\ref{fig:supp_qual_megascene100}, \ref{fig:supp_qual_realestate10k} and \ref{fig:supp_qual_nerfllff} on complex camera motion, including dolly zoom-in, zoom-out and wavy paths. Some failure cases are also shown. Figure~\ref{fig:supp_qual_nerfllff} also provides application-oriented qualitative results on stylized scenes. Style-NeRF2NeRF~\citep{fujiwara2024style} generates the stylized inputs, and the same camera-controlled generation pipeline is then applied.

\end{document}